\documentclass[12pt]{article}
\usepackage[pdftex]{graphicx}

\usepackage[preprint,nonatbib]{neurips_2022}

\usepackage[utf8]{inputenc} 
\usepackage[T1]{fontenc}    
\usepackage{url}            
\usepackage{amsfonts}       
\usepackage{nicefrac}       
\usepackage{microtype}      %
\usepackage{fancyhdr}       
\usepackage{graphicx}       
\usepackage{amsthm}
\usepackage{amsmath}
\usepackage{booktabs, multirow} 
\usepackage{soul}
\usepackage[table]{xcolor} 
\usepackage{changepage,threeparttable} 

\usepackage{mathtools} 
\usepackage{tikz} 
\usepackage{graphicx,multicol}
\usepackage{caption}
\usepackage{subcaption}
\usepackage{multirow}
\usepackage{enumitem}

\usepackage[ruled,linesnumbered]{algorithm2e}
\SetKwComment{Comment}{/* }{ */}

\theoremstyle{definition}
\newtheorem{definition}{Definition}[section]
\newcommand{\para}[1]{{\noindent \textbf{#1}}}

\newtheorem{lemma}{Lemma}

\newcommand{\ourapp}{FedAUC}
\newcommand{\ourapprr}{$\text{FedAUC}_{\text{RR}}$}
\newcommand{\ourapplap}{$\text{FedAUC}_{\text{Lap}}$}

\newcommand{\locallap}{LocalLaplace}
\newcommand{\globallap}{GocalLaplace}

\title{Differentially Private AUC Computation \\ in Vertical Federated Learning}

\author{%
  Jiankai Sun\thanks{Bytedance Inc. Corresponds to \texttt{\{jiankai.sun, chong.wang\}@bytedance.com}} 
   \And
 Xin Yang
 \And 
 Yuanshun Yao 
 \And 
 Junyuan Xie
 \And 
 Di Wu
 \And 
 Chong Wang
}

\begin{document}

\maketitle

\begin{abstract}
Federated learning has gained great attention recently as a privacy-enhancing tool to jointly train a machine learning model by multiple parties. As a sub-category, vertical federated learning (vFL) focuses on the scenario where features and labels are split into different parties. The prior work on vFL has mostly studied how to protect label privacy during model training. However, model evaluation in vFL might also lead to potential leakage of private label information. One mitigation strategy is to apply label differential privacy (DP) but it gives bad estimations of the true (non-private) metrics. In this work, we propose two evaluation algorithms that can more accurately compute the widely used AUC (area under curve) metric when using label DP in vFL. Through extensive experiments, we show our algorithms can achieve  more accurate AUCs compared to the baselines.

\end{abstract}

\section{Introduction}
\label{sec:intro}

With increasing concerns over data privacy in machine learning, regulations like CCPA\footnote{California Consumer Privacy Act}, HIPAA\footnote{Health Insurance Portability and Accountability Act}, and GDPR\footnote{General Data Protection Regulation, European Union} have been introduced to regulate how data can be transmitted and used. 
To address privacy concerns, \textit{federated learning}~\cite{mmr+17,hhh+20,ym20,gcyr20} has become an increasingly popular tool to enhance privacy by allowing training models without directly sharing their data. As a specific type of federated learning, \textit{vertical federated learning} (vFL)~\cite{yang2019federated,gupta2018distributed,vepakomma2018split} focuses on the application when the data is separately owned by multiple parties and each party holds either features or labels of the same data subjects. For example, in online advertising~\cite{li2021label,sunDefending2021}, advertisers and the advertising platform (e.g. Google, Meta, and TikTok) can jointly train a model to predict the conversion rate (CVR) of an ad impression. In vFL protocol, the model is split layerwise and disjointly owned by advertisers and the platform, and the training and test are done by exchanging backward gradients and forward embeddings rather than the raw data. Therefore private information like user information (i.e. features) remains on the platform while the conversion events (i.e. labels) are kept to advertisers. We refer to the platform which only owns raw features and is in charge of the communication as the \textit{server}, and the individual parties (advertisers) which own private labels as \textit{clients}.

Despite the practical usefulness of vFL, researchers have shown that vFL can still leak data information indirectly. For example, Li et al~\cite{li2021label} demonstrate that the gradient updates between the server and client can leak label information. However, the prior work on vFL privacy mostly focuses on the model training and there can also be privacy leaked from model evaluation. Specifically, the private label information owned by clients can be leaked to the server when computing evaluation metrics. For example, Stoddard el al.~\cite{Dp4ML2014} show that the server can simply guess the labels, plot the ROC curve, and compare it with the given ROC curve.

In this work, we focus on protecting label privacy when multiple parties jointly evaluate the trained vFL model. Specifically, we want to protect privacy when computing AUC. AUC is a widely used and standard metric; therefore it is a must when evaluating models. However, computing AUC in vFL while protecting label information is challenging because AUC requires sorting the testing samples by predicted score globally (i.e. from all clients) which is a procedure more complicated than other simpler metrics (e.g. test loss or accuracy). For example, the server can compute global test loss easily by simply asking all clients to send their local loss and then compute the global test loss as the weighted sum (weighted by the number of local samples). However, simply aggregating local AUC does not lead to the correct global AUC because the overall AUC requires global ranking.


To address the challenge, we leverage probabilistic interpretation of AUC related to Mann-Whitney U statistics~\cite{AUCProb2002,dpforclassifierEvaluation}. We view AUC as the probability of correct ranking of a random positive-negative pair and we can avoid requiring clients to send their labels to the server. Our goal is to achieve \textit{label differential privacy}~\cite{labeldp21} since the class label is often the most sensitive information in a prediction task. To this end, we propose two DP mechanisms. The first leverages randomized response to flip local labels in the clients, and then we debias the AUC computed with flipped labels. The second approach adopts the Laplace mechanism (Gaussian mechanism is applicable too) to add noise to the shared intermediate information between the server and clients to calculate the AUC. We conduct extensive experiments to demonstrate the effectiveness of our proposed approaches.

\section{Preliminaries}
\label{sec:preliminaries}

We start with introducing some background knowledge of our work.

\subsection{Vertical Federated Learning} 

\para{Training.} The training of vFL includes forward pass and backward gradients computation. During the forward pass, the party without labels (server) sends the intermediate layer (\textit{cut layer}) outputs rather than the raw data to the party with labels (clients), and the clients completes the rest of the forward computation to obtain the training loss. In the backward phase, to compute the gradients w.r.t model parameters, the client performs backpropagation from its training loss, computes its own parameters' gradients, and updates its own model. The client also computes the gradients w.r.t cut layer outputs and sends it to the server, and then server uses the chain rule to compute gradients of its model parameters, and update its model.

\para{Inference.} During inference time, the server computes the cut layer embedding and sends it to the clients. Clients then execute the rest of forward computation to compute the prediction probability. 

We focus on the setting of multi-party vFL which contains one server and multi clients. The labels are distributed in multi clients and our proposed approaches can compute the evaluation metric AUC with label differential privacy. 

\subsection{Label Differential Privacy}



In our work, we focus on protecting the privacy of label information. Following ~\cite{labeldp21}, we define label differential privacy as the following:

\begin{definition}[Label Differential Privacy]
Let $\epsilon,\delta\in \mathbb{R}_{\geq 0}$, a randomized mechanism $\mathcal{M}$ is $(\epsilon,\delta)$-label differentially private (i.e. $(\epsilon,\delta)$-LabelDP), if for any of two neighboring training datasets $D,D'$ \textit{that differ in the label of a single example},
and for any subset $S$ of possible output of $\mathcal{M}$, we have

\begin{align*}
    \Pr[\mathcal{M}(D)\in S]\leq e^{\epsilon}\cdot \Pr[\mathcal{M}(D')\in S]+\delta.
\end{align*}
\end{definition}

If $\delta = 0$,  then $M$ is $\epsilon$-label differentially private (i.e. $\epsilon$-LabelDP).

Our proposed approaches also shares the same setting with local DP ~\cite{localdpDuchi2013,localdprappor2014,localdp2008,localdp2019} which assumes that the data collector (server in our paper) is untrusted. Following the same setting with local DP, in our proposed approaches, each client locally perturbs their private labels with a DP mechanism and transfers the perturbed version to the  server. After receiving all clients' perturbed data, the server calculates the statistics and publishes the  result of AUC. We define local DP as the following:

\begin{definition}[Local Differential Privacy]
Let $\epsilon > 0$ and $ 1 > \delta \geq 0$, a randomized mechanism $\mathcal{M}$ is $(\epsilon,\delta)$-local differentially private (i.e. $(\epsilon, \delta)$-LocalDP), if and only if for any pair of input values $v$ and $v'$ in domain $D$, and for any subset $S$ of possible output of $\mathcal{M}$, we have

\begin{align*}
    \Pr[\mathcal{M}(v)\in S]\leq e^{\epsilon}\cdot \Pr[\mathcal{M}(v')\in S]+\delta.
\end{align*}
\end{definition}

If $\delta = 0$,  then $M$ is $\epsilon$-local differentially private (i.e. $\epsilon$-LocalDP).

\begin{definition}[Sensitivity]
Let $d$ be a positive integer, $\mathcal{D}$ be a collection of datasets, and $f: \mathcal{D} \rightarrow \mathcal{R}^d$ be a function. The sensitivity  of a function, denoted $\Delta f$, is defined by $\Delta f = max ||f(D)-f(D')||_p
$ where the maximum is over all pairs of datasets $D$ and $D'$ in $\mathcal{D}$ differing in at most one element and  $||\cdot||_p
$ denotes the $l_p$ norm.

\end{definition}




\begin{lemma}[Laplace Mechanism]  Laplace mechanism preserves $(\epsilon, 0)$-differential privacy if the random noise is drawn from $Lap(\Delta/\epsilon)$.
\end{lemma}

\subsection{ROC Curve and AUC}

 In a binary classification problem, given a threshold $\theta$, a predicted score $s_i$ is predicted to be $1$ if $s_i \ge \theta$. Given the ground-truth label and the predicted label (at a given threshold $\theta$), we can compute True Positive Rate (TPR) and False Positive Rate (FPR). TPR (i.e. recall) is defined as $TPR(\theta) = \frac{TP(\theta)}{TP(\theta) + FN(\theta)}$ and False Positive Rate (FPR) is defined as $FPR(\theta) = \frac{FP(\theta)}{FP(\theta) + TN(\theta)}$. Receiver operating characteristic (ROC) curve plots TPR (x-axis) vs. FPR (y-axis) over all possible thresholds $\theta$, and AUC is the area under the ROC curve. A perfect classifier has AUC $1.0$ while a classifier giving random predictions has AUC $0.5$.

\para{Privacy Leakage in AUC.} Researchers have shown AUC computation can cause privacy leakage. Matthews and Harel~\cite{Roc2013} demonstrate that by using a subset of the ground-truth data and the computed ROC curve, the data underlying the ROC curve can be reproduced accurately. Stoddard el al.~\cite{Dp4ML2014} show that an attacker can determine the unknown label by simply enumerating over all labels, guessing the labels, and then checking  which guesses lead to the given ROC curve. They propose a differentially private ROC curve computation algorithm that adds DP noise to TPRs and FPRs.



\subsection{Probabilistic Interpretation of AUC}


Area under ROC curve (AUC) can be related to Mann-Whitney U statistics ~\cite{AUCProb2002,dpforclassifierEvaluation} by 
 viewing it as the probability of correct ranking of a random positive-negative pair. Suppose we have $M$ samples with $P$ positive and $N$ negative samples, where $M = P + N$. Given a classifier $\mathcal{F}$ that outputs a prediction score $s_i$ for each sample with index $i$ from 1 to $M$. To compute AUC, we first sort all samples based on their prediction scores with an increasing order, i.e. each predicted score $s_i$ is assigned a rank $r_i$ where $r_i = 0$ indicates sample with index $i$ has the lowest prediction score $s_i$ and where $r_i = M-1$ indicates the highest. Then AUC can be computed as the following:
 

\begin{equation}
\label{eq:auc_ranking}
    \text{AUC} = \frac{\sum_{i=1}^{M} r_i \cdot y_i - \frac{P(P-1)}{2} }{PN}
\end{equation}

where $y_i \in \{0, 1\}$ is the ground-truth label for sample $i$. 
We include more details in Section ~\ref{sec:auc_probabilistic_view} of the Appendix.

\section{Threat Model}
\label{sec:threat}

We formally define our scenario. In our vFL setting, there are multiple label parties (i.e. clients) that own private labels (i.e. Y)\footnote{they can also own some features} and there is a central non-label party (i.e. server) that owns features and is responsible for computing global AUC from all clients. The model (early layers) owned by the server is shared by all clients, and each client owns its individual model (late layers) trained on their own private data. Before the training starts, the server and each client split their data into training and test set. The model is trained using the normal vFL protocol.


Our work focuses on the evaluation time and the goal of the server is to compute global AUC without letting clients directly share their private test data. In other words, clients cannot directly send the test data (i.e. private labels) to the server for it to compute AUC. Specifically, we are interested in protecting label information and therefore it is required that the exchanged information between client and server excludes the ground-truth test labels (Y). 


Note that the exchanged information might contain some indirect information that can be used by the server to infer clients' test labels (Y), e.g. the predicted score $f(X)$, which we assume to be considered non-sensitive by nature of the task. For example, in online advertising, clients (i.e. advertisers) have to send the predicted scores, i.e. predicted conversion rate (CVR), to the server (i.e. advertising platforms like Google, Meta, or Tiktok) so that the server can compute the bid price~\cite{bidding2018}. In this case, the predicted scores have to be shared with the server in order to perform the task, and clients have already agreed to share them before participating in vFL. We include more discussion on the privacy sensitivity of predicted scores in Section~\ref{sec:sorting_module} in the Appendix. 



\section{Proposed Algorithm 1: Randomized Response Mechanism}
\label{sec:methods}


We introduce our first algorithm to compute AUC when applying label DP to vFL for evaluation. The algorithm is based on randomized response~\cite{RR} and we name it as \ourapprr{}. We include the workflow in Figure ~\ref{fig:fedauc_illustration_rr} in the Appendix. We now explain the algorithm step by step.


\subsection{Step 1: Clients Flip Their Local Labels}

Randomized response (RR) is $\epsilon$-LabelDP and  works as following: let $\epsilon$ be a parameter and let $y \in \{0, 1\}$ be the true label. Given a query of $y$, RR will respond with a random draw $\tilde{y}$ from the following probability distribution: 


\begin{equation}
\label{eq:rr}
    \Pr[\tilde y = \hat y] = \begin{cases}
     \frac{e^\epsilon}{1+e^\epsilon} & \text{for $y = \hat y$,} \\
    \frac{1}{1+e^\epsilon} & \text{otherwise.}
                            \end{cases}
\end{equation}




Clients can leverage randomized response (Algorithm ~\ref{alg:clients_flipping_labels} in the Appendix) to flip their owning labels as a preprocessing step before computing the AUC. It's worth mentioning that all labels are only flipped once and the generated noisy labels can then be used for further evaluations multi-times. 


\subsection{Step 2: Server Computes AUC from Flipped Labels}
\label{sec:corr_auc}

We now talk about how to compute the AUC with the flipped labels (detailed description can be seen in Algorithm ~\ref{alg:clients_server_cal_auc} in the Appendix). 
 Since the corresponding AUC is computed with flipped labels,
 we denote this AUC as \emph{noisy AUC}: $\text{AUC}^{D_\text{noisy}}$. It has four steps:

\begin{enumerate}[leftmargin=*]
    \item \textbf{Clients Execute.} Suppose we have $K$ clients. Each client $C_k$ computes the prediction scores $s_k = f(X_k)$ for all its owning data points and sends shuffled $s^k$ to the server.  
    \item \textbf{Server Executes.} The server aggregates all the prediction scores and sort them in an increasing order. Each prediction score $s_i^k$ ($i \in [1, M]$) will be assigned a ranking order $r_i^k$. The instance with the highest order will be assigned $r = M - 1$, and the second highest one will be assigned $M -2$ and so on. The smallest ranking score is $r = 0$. The server sends each ranking order $r_i^k$ back to the corresponding client $C_k$ which owns $s_i^k$. 
    \item \textbf{Clients Execute.} Each client $C_k$ then aggregates all its received ranking orders with the noisy label generated in Algorithm ~\ref{alg:clients_flipping_labels}. The corresponding result is $\text{localSum}_k = \sum_{i}^{|{Y_k}|} r_i^k \cdot {y_i}'$, where $y_i'$ is the flipped version of ground-truth label $y_i$. The client also computes its number of noisy positive ($\text{localP}$) and negative ($\text{localN}$) instances respectively. Here $\text{localP} = \sum_{i}^{|{Y_k}|} {y_i}'$ and  $\text{localN} = \sum_{i}^{|{Y_k}|} (1- {y_i}') $.
    Each client $C_k$ sends its $\text{localSum}_k$, $\text{localP}_k$, and $\text{localN}_k$ to the server. 
    \item \textbf{Server Executes.} After receiving all clients' $\text{localSum}$, $\text{localP}$ and $\text{localN}$, the server aggregates them to get the corresponding global values. Here $\text{globalSum} = \sum_{k}^K \text{localSum}_k$, $\bar{P} = \sum_{k}^K \text{localP}_k$, and $\bar{N} = \sum_{k}^K \text{localN}_k$. The server then leverages equation ~\eqref{eq:auc_ranking} to compute the AUC. We then get the corresponding noisy AUC: $\text{AUC}^{D_{\text{noisy}}} = \frac{{\text{globalSum} - {\bar{P}(\bar{P}-1)}/{2}}}{{\bar{P}\bar{N}}}$.
\end{enumerate}

It's possible that the clients can send their flipped labels and corresponding prediction scores to the server directly and the server then computes $\text{AUC}^{D_{\text{noisy}}}$ in a centralized way. Both methods will get the same value of the noisy AUC. To be consistent with our other mechanisms, we instead introduce the one with four steps here.

\subsection{Step 3: Server Debiases AUC}
\label{sec:covert_auc_to_real}

Section ~\ref{sec:corr_auc} shows how the server compute $\text{AUC}^{D_{\text{noisy}}}$ with using clients' local flipped labels. Empirically, the noisy AUC can be useless with a small privacy budget $\epsilon$ (see experimental results in Table ~\ref{tab:noisy_auc_randomized_response_wo_converting} in the Appendix). In this section, we show how to covert  $\text{AUC}^{D_{\text{noisy}}}$ to the final AUC that we are interested in and we name it as $\text{AUC}^{D_{\text{clean}}}$. 
$\text{AUC}^{D_{\text{clean}}}$ is a more accurate estimation of the ground-truth AUC than $\text{AUC}^{D_{\text{noisy}}}$.

\cite{ConvertAUC2015} proposed to learn from noisy binary labels via class-probability estimation. They draw $(X, Y) \sim D$. The instance $X$ is unchanged: however, the label is altered such that positive samples have labels flipped with probability $\rho_{+}$, while negative samples have labels flipped with probability $\rho_{-}$. 
We note the AUC computed based on the ground-truth labels as $\text{AUC}^{D_{\text{clean}}}$. \cite{ConvertAUC2015} proved the following relation between $\text{AUC}^{D_{\text{noisy}}}$ and $\text{AUC}^{D_{\text{clean}}}$:

\begin{equation}
\label{equ:auc_convertion}
  \text{AUC}^{D_{\text{clean}}}  =  \frac{\text{AUC}^{D_{\text{noisy}}} - \frac{\alpha+\beta}{2}}{1-\alpha-\beta}    
\end{equation}

where 
\begin{equation}
    \label{equ:alpha_beta}
    \alpha = \frac{(1-\pi) \rho_{-}}{\pi(1-\rho_{+}) + (1-\pi)\rho_{-}}, \quad 
    \beta = \frac{\pi \rho_{+}}{\pi \rho_{+} + (1-\pi)(1-\rho_{-})}
\end{equation}

Here $\pi = P_{D}(Y=1)$ is the positive ratio (base rate) in the clean data $D$. $\pi$ is unknown to the server. However, the server can estimate $\pi$ from the known information of the noisy data $D'$. Suppose we observe $\bar{M}$ positive examples and $\bar{N}$ negative examples in the noisy data $D'$. We then estimate $\pi$ as $\pi' = \frac{P'}{P' + N'}$ where $P'$ and $N'$ are estimated positive and negative numbers in the clean data $D$. And they can be estimated as $P' = \frac{\bar{P}(1-\rho_{-}) - \bar{N}\rho_{-}}{1-\rho_{+}-\rho_{-}}$, and $\quad N'= \bar{P}+\bar{N}-P'$. More details can be in Section ~\ref{sec:estimate_pi} of the Appendix.


After we get get $\text{AUC}^{D_{\text{noisy}}}$ and $\pi'$, we can leverage equation ~\eqref{equ:alpha_beta} to compute $\alpha$ and $\beta$, and then use equation ~\ref{equ:auc_convertion} to get the $\text{AUC}^{D_{\text{clean}}}$ that we are interested in. Algorithm ~\ref{alg:real_auc_from_corr} in the Appendix shows how the server compute $\text{AUC}^{D_{\text{clean}}}$ from  $\text{AUC}^{D_{\text{noisy}}}$, $\bar{P}$, $\bar{N}$, and Label DP budget $\epsilon$.

\para{Privacy and Utility Analysis.}
Our proposed approach is $\epsilon$-LabelDP since all of operations are on the noisy labels flipped by randomized response which is $\epsilon$-LabelDP. We include the utility (variance of computed AUC) analysis in Section~\ref{sec:utility_rr} of the Appendix.




\section{Proposed Algorithm 2: Laplace and Gaussian Mechanism}
\label{sec:method_2}

In this section, we introduce how to compute the AUC with label differential privacy by leveraging Laplace and Gaussian mechanism. Here we use Laplace mechanism as an example. Other settings for the Gaussian mechanism will be the same. 

\subsection{Overall Workflow}
The workflow of this method is as shown in Figure ~\ref{fig:fedauc_illustration_lap} in the appendix and the detailed  Algorithm is shown in Algorithm ~\ref{alg:clients_server_cal_auc_2} in the Appendix. The algorithm has four steps (the first two steps are the same as \ourapprr{}):
 
\begin{enumerate}[leftmargin=*]
    \item \textbf{Clients Execute.} Each client $C_k$ computes the prediction scores $s_k = f(X_k)$ for all its owning data points and sends shuffled $s^k$ to the server.  
    \item \textbf{Server Executes.} The server aggregates all the prediction scores and sort them in an increasing order. Each prediction score $s_i^k$ ($i \in [1, M]$) will be assigned a ranking order $r_i^k$. 
    The server sends each ranking order $r_i^k$ back to the corresponding client $C_k$ which owns $s_i^k$. 
    \item \textbf{Clients Execute.} Each client $C_k$ then aggregates all its received ranking orders with its label. The corresponding result is $\text{localSum}_k = \sum_{i}^{|{Y_k}|} r_i^k \cdot {y_i}$, where $y_i$ is the ground-truth label. The client also computes its number of  positive ($\text{localP}$) and negative ($\text{localN}$) instances respectively. Here $\text{localP} = \sum_{i}^{|{Y_k}|} {y_i}$ and  $\text{localN} = \sum_{i}^{|{Y_k}|} (1- {y_i}) $. Each client $C_k$ adds noise to $\text{localSum}_k$ and $\text{localP}_k$ to get  $\text{localSum}_k'$ and $\text{localP}_k'$ respectively. The client $C_k$ sends $\text{localSum}_k'$, $\text{localP}_k'$, and  $\text{localN}_k'$ ($\text{localN}_k' = |Y_k| - \text{localP}_k'$) to the  server. We talk about how to add noises to these local statistics in Section ~\ref{sec:adding_noise_to_local_statistics}.
    \item \textbf{Server Executes.} After receiving all clients' $\text{localSum}'$, $\text{localP}'$ and $\text{localN}'$, the server aggregates them to get the corresponding global values. Here $\text{globalSum} = \sum_{k}^K \text{localSum}_k'$, $\bar{P} = \sum_{k}^K \text{localP}_k'$, and $\bar{N} = \sum_{k}^K \text{localN}_k'$. The server then leverages equation ~\eqref{eq:auc_ranking} to compute the AUC. The final AUC is: $\text{AUC} = \frac{{\text{globalSum} - {\bar{P}(\bar{P}-1)}/{2}}}{{\bar{P}\bar{N}}}$.
\end{enumerate}

\subsection{Adding Noise to Local Statistics}
\label{sec:adding_noise_to_local_statistics}

We explain in details how to perturb $\text{localSum}$ and $\text{localP}$ for each client in the algorithm. Both Gaussian and Laplace mechanisms can be leveraged to generate the corresponding DP noise. Without loss of generality, we use Laplace as an example. Laplace mechanism preserves $(\epsilon, 0)$-differential privacy if the random noise is drawn from $Lap(\Delta/\epsilon)$ where $\Delta$ is the $l_1$ sensitivity. We name this method as \ourapplap{}. The noise is added as the following:

\begin{enumerate}[leftmargin=*]
    \item Adding noise to $\text{localSum}$: Each client $C_k$ decides its own sensitivity by selecting its maximum ranking order as $\Delta_{\text{localSum}^k} = \max_{i \in [1, |Y_k|]}(r_i^k)$. Given a privacy budget $\epsilon_{localSum}$, client $C_k$ draws the random noise from $\textsf{Lap}(\Delta_{\text{localSum}^k}/\epsilon_{localSum})$.
    \item Adding noise to $\text{localP}$: Each client $C_k$ sets $\Delta_{\text{localP}^k} = 1$. Given a privacy budget $\epsilon_{localP}$, client $C_k$ draws the random noise from $\textsf{Lap}(1/\epsilon_{localP})$.
\end{enumerate}

\para{Privacy Analysis.}
The total privacy budget $\epsilon =\epsilon_{localSum} + \epsilon_{localP}$. We use a parameter $\alpha$ to control the budget allocation: $\epsilon_{localSum} = \alpha \epsilon$ and $\epsilon_{localP} = (1- \alpha) \epsilon$. Next we explain a a better allocation than using $\alpha$.

\subsection{Adaptive Allocation of DP Budget} 
\label{sec:adaptive_allocation_of_dp_budget} 

Finding $\alpha$ that achieves a good tradeoff between utility and privacy can be difficult and would require many trials and errors if using cross validation. Therefore we propose an algorithm that automatically determines the allocation of the DP budget adaptively on each client.


Our inspiration comes from the special case where each client only owns 1 sample, i.e., $|Y_k|=1$ for each client $C_k$.
In this case, 
it turns out that $\text{localSum}_k = r^k \cdot \text{localP}_k$. Since $r^k$ is public, we do not need to generate two independent random variables;
we can sample $s\sim \mathsf{Lap}(1/\epsilon)$,
and report $\text{localP}_k':=\text{localP}_k+s$ and $\text{localSum}_k':=r^k\cdot \text{localP}_k'$.

The above example shows that when $\text{localSum}_k$ and $\text{localP}_k$ are highly correlated, then the privacy budget can be allocated more efficiently.
Based on this observation, we can analyze the general case.
Fix the client $C_k$.
Let $\mathbf{y}:=(y_1,\cdots,y_{|Y_k|})\in \{0,1\}^{|Y_k|}$ be the labels held by $C_k$,
and $\mathbf{r}:=(r_1^k,\cdots,r_{|Y_k|}^k)\in \mathbb{Z}^{|Y_k|}$ be their rankings.
Then we have $\text{localP}_k=\langle \mathbf{u},\mathbf{y}\rangle$ and $\text{localSum}_k = \langle \mathbf{r},\mathbf{y}\rangle$,
where $\mathbf{u}:=(1,1,\cdots,1)$ is the all-one vector.

When $\mathbf{u}$ and $\mathbf{r}$ are parallel, namely $\mathbf{r} = r\cdot \mathbf{u}$ for some $r\in \mathbb{R}$,
we can still allocate all the budget on $\text{localP}_k'$,
and report $\text{localSum}_k'=r\cdot \text{localP}_k'$.
Otherwise,
we can decompose $\mathbf{r}$ into two orthogonal components:
we can write $\mathbf{r}:=\mathbf{u}'+\mathbf{v}'$ where $\mathbf{u}'\parallel \mathbf{u}$ and $\mathbf{u}'\perp \mathbf{v'}$.
Then we have 
\begin{align*}
    \text{localSum}_k=\langle \mathbf{r},\mathbf{y}\rangle = \langle \mathbf{u'},\mathbf{y}\rangle+\langle \mathbf{v'},\mathbf{y}\rangle.
\end{align*}
In order to make $(\text{localP}_k,\text{localSum}_k)$ $\epsilon$-DP,
by the post-processing property of differential privacy,
it is sufficient to have $(\langle \mathbf{u}',y\rangle, \langle \mathbf{v}',y\rangle)$ to be $\epsilon$-DP. We can use a parameter $\beta$ to control the budget allocation so that $\langle \mathbf{u}',y\rangle$ achieves $\beta \epsilon$-DP and $\langle \mathbf{v}',y\rangle$ achieves $(1-\beta)\epsilon$-DP.
If we use Laplace mechanism, this means we sample $s_1\sim \mathsf{Lap}(1/(\beta \epsilon))$ and $s_2\sim \mathsf{Lap}(1/((1-\beta)\epsilon))$,
and report $\text{localP}_k':=\langle \mathbf{u},y\rangle+s_1$,
$\text{localSum}_k':=\langle \mathbf{u}',y\rangle+\|\mathbf{u}'\|_{\infty}\cdot s_1+\langle \mathbf{v}',y\rangle+\|\mathbf{v}'\|_{\infty}\cdot s_2$.

We can tune the parameter $\beta$ to achieve small standard deviation of estimated AUC. However, the server may need additional validation data to select a optimal $\beta$ which performs the best. Sometimes, it is impractical in the setting of FL. We now show how to adaptively choose the allocation parameter $\beta$ for each client.
Simply, we use the heuristic of choosing $\beta$ that minimizes the variance of $\text{localSum}_k'$ for client $k$,
which is equivalent to minimizing the variance of $\|\mathbf{u}'\|_{\infty}\cdot s_1+\|\mathbf{v}'\|_{\infty}\cdot s_2$.

\section{Experiments}
\label{sec:experiments}

In this section, we show the experimental results of evaluating our proposed approaches. 

\subsection{Experimental Setup}

\textbf{Dataset.}
We evaluate the proposed approaches on Criteo~\footnote{{https://www.kaggle.com/c/criteo-display-ad-challenge/data}}, which is a large-scale industrial binary classification dataset (with with approximately $45$ million user click records) for conversion prediction tasks. 
We computed the AUC on the test set which contains $M=458,407$ where $P = 117,317$ and $N = 341,090$ for $3$ epochs.

\textbf{Model.}
 We modified a popular deep learning model architecture WDL \cite{cheng2016wide} for online advertising. 
 Note that our goal is not to train the model that can beat the state-of-the-art, but to test the effectiveness of our proposed federated AUC computation approach.

\textbf{Ground-truth AUC.}
We compare our proposed \ourapp{} with two AUC computation libraries (their computed results work as ground-truth and have no privacy guarantee): 1) scikit-learn\footnote{\url{https://scikit-learn.org/stable/modules/generated/sklearn.metrics.auc.html}}; 2) Tensorflow \footnote{\url{https://www.tensorflow.org/api_docs/python/tf/keras/metrics/AUC}}. Both approximate the AUC (Area under the curve) of the ROC. In our experiments, we set $\text{num\_thresholds}=1,000$ for Tensorflow. We use the default values for other parameters.

\textbf{Evaluation Metric.} For each method, we run the same setting for $100$ times (change the random seed every time) and use the corresponding standard deviation of the computed AUC as our evaluation metric\footnote{We only include the standard deviation of computed AUCs rather than the mean because we find that the mean of our AUCs is close to the ground-truth AUC. See the results of mean AUC in Table~\ref{tab:auc_std_criteo} and ~\ref{tab:auc_std_criteo_large} in the Appendix.}. A good computation method should achieve a small std of the computed AUC. 



\subsection{Experimental Results}

\subsubsection{Results of  \ourapprr{}}

\textbf{Effectiveness of \ourapprr{}}. We report the standard deviation of the AUC estimated by \ourapprr{} with different $\epsilon$ in Table ~\ref{tab:std_auc_rr_global_local_adaptive}.  We put more detailed results in Table ~\ref{tab:auc_std_criteo} and ~\ref{tab:std_auc_w_noisy_P_and_N} in the Appendix. We can observe that vanilla \ourapprr{} (with $\epsilon = + \infty$ and $\rho_{+} = \rho_{-}=0$) which has no DP guarantee can achieve almost the same value as the ones computed by scikit-learn and Tensorflow. The corresponding results demonstrated the correctness and effectiveness of \ourapprr{}. Results in Table ~\ref{tab:noisy_auc_randomized_response_wo_converting} in the Appendix shows the results without  debiasing as we described in Section ~\ref{sec:covert_auc_to_real}, which demonstrates that converting the noisy AUC to the clean AUC is badly needed.





\textbf{Impact of data size.} We also tested the effectiveness of \ourapprr{} on a larger evaluation set which has 4,584,062 data samples which is 10 times larger than the small one. The detailed statistics of these two dataset can be seen in Table ~\ref{tab:evaluation_set_statistics} in the Appendix. As shown in Table ~\ref{tab:auc_std_criteo_large} (in the Appendix) and Figure ~\ref{fig:auc_dataset_size_alpha_criteo} (a), the AUC on the large dataset is approximately $1/\sqrt{10}$ of the value achieved on the small dataset with the same privacy budget $\epsilon$. It indicates that increasing the dataset size can reduce the standard deviation of the estimated AUC. The same conclusion is applicable to \ourapplap{} too.

\begin{figure*}[ht!]
\captionsetup[subfigure]{labelformat=empty}
  \begin{subfigure}{0.3\linewidth}
  \centering
    \includegraphics[width=1.0\linewidth]{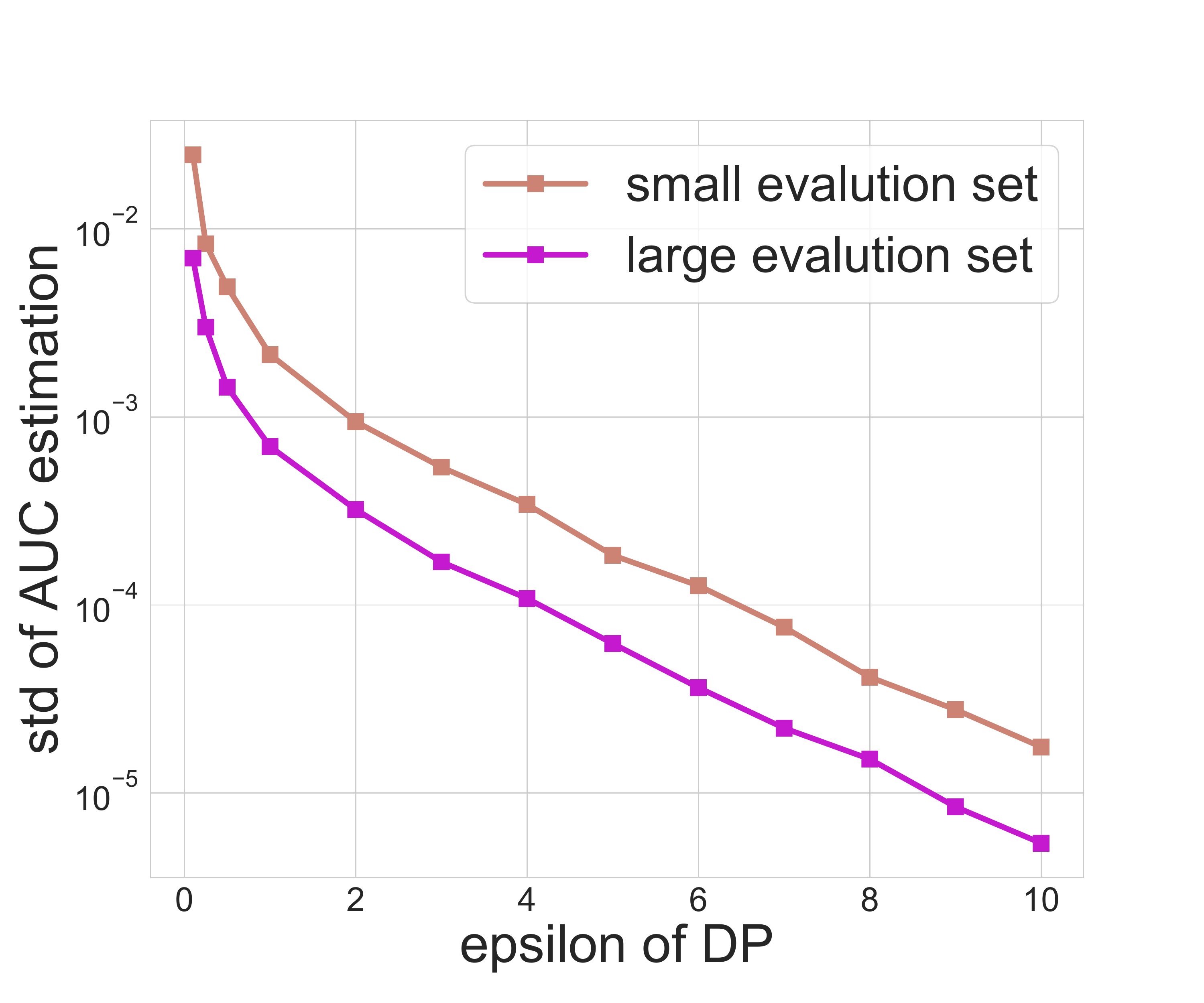}
      \caption{(a): standard deviation of AUC estimated by \ourapprr{} on two evaluation sets with different sizes. }
  \end{subfigure}
  \hfill
  \begin{subfigure}{0.3\linewidth}
  \centering
    \includegraphics[width=1.0\linewidth]{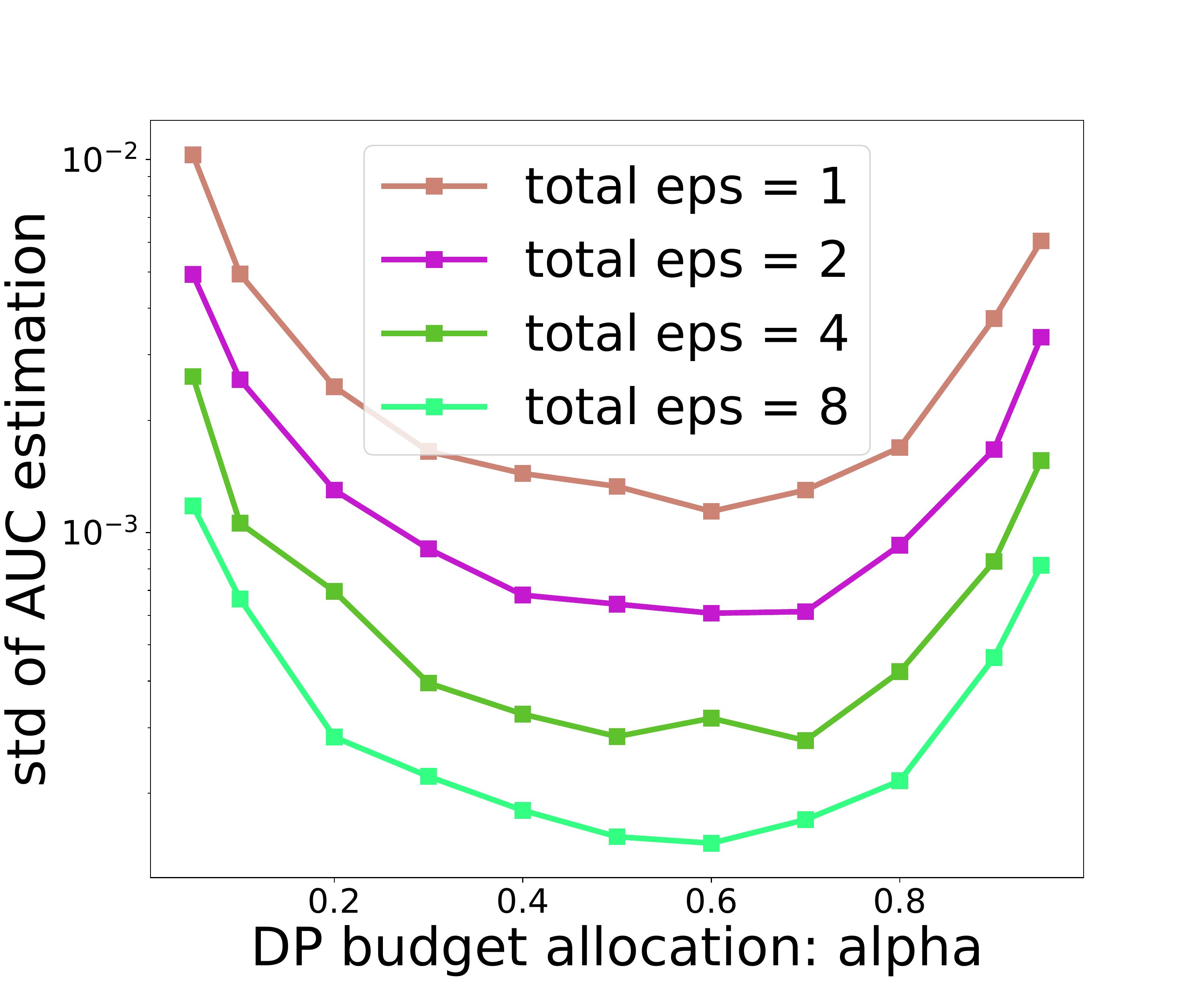}
    \caption{(b): Sensitivity of $\alpha$: standard deviation of AUC estimated by \ourapplap{} with different $\epsilon$ on the IID setting.}
  \end{subfigure}
 \hfill
\begin{subfigure}{0.3\linewidth}
  \centering
    \includegraphics[width=1.0\linewidth]{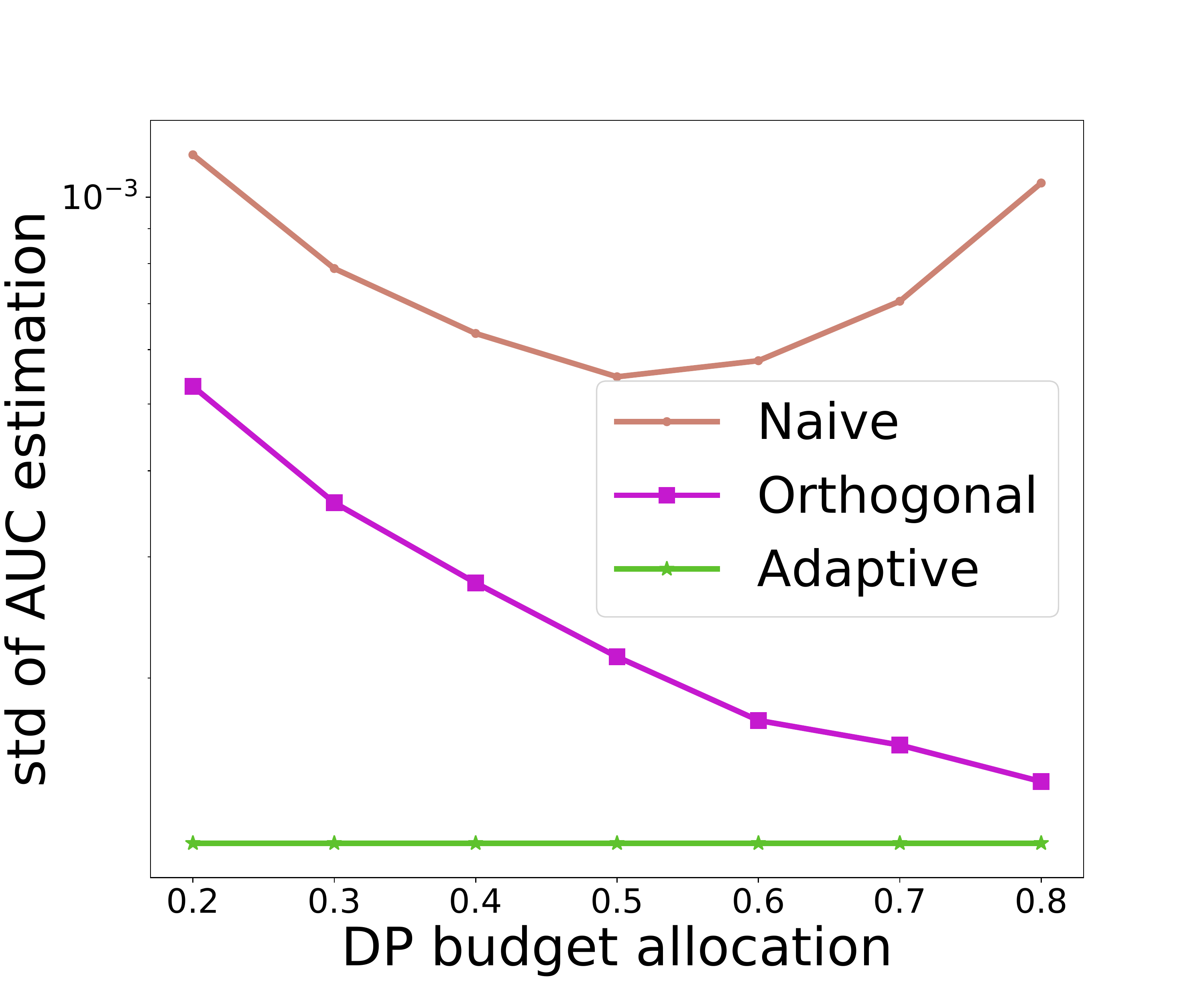}
    \caption{(c): Advantage of adaptive allocation: standard deviation of AUC estimated by \ourapplap{} with different allocation methods on the Non-IID setting.}
  \end{subfigure}
  
   \caption{ a) effects of data size; b) sensitivity of $\alpha$; c) advantage of  adaptive allocation.}
 \label{fig:auc_dataset_size_alpha_criteo} 
 \end{figure*}


 
\subsubsection{Results of  \ourapplap{}}

 Both Gaussian and Laplace mechanism can be use in \ourapplap{}. In our experiments, we adopt Laplace mechanism as an example to show the effectiveness of our proposed technique. We refer this technique as \locallap{}. We also add a comparison partner \globallap{} here. All clients use $M - 1$ as the sensitivity when adding noise to their local statistics $\text{localSum}$. Given a privacy budget $\epsilon_{localSum}$, client $C_k$ draws the random noise from $Lap((M-1)/\epsilon_{localSum})$. Each client adds the same amount of noise to its $\text{localSum}$. $\epsilon_{localP}$ is kept the same as \locallap{}. 
 


Unlike \ourapprr{}, 
\ourapplap{} is sensitive to the number of clients and number of samples per client has. 
Based on the setting of how to assign data sample to clients, we provide two simulations to conduct the corresponding experiments.

\begin{enumerate}[leftmargin=*]
    \item IID: all data points are uniformly assigned to the clients. 
    \item Non-IID: all data points are assigned to the clients based on their corresponding prediction scores. Data samples with similar prediction scores will be assigned to the same client. 
\end{enumerate}

We divide $M$ data samples into $K$ clients based on the IID and Non-IID settings and each client has $\frac{M}{K}$ data samples on average. We also provide an extreme scenario to simply our analysis: each client only has one data sample ($M = K$). Under this  setting and suppose we use accurate $P$ and $N$ to compute the AUC, we can conclude from Equation \ref{eq:var_globallap}  and \ref{eq:var_locallap} (both in the Appendix) that: $\frac{\text{AUC}_{\text{globalLaplace}}}{\text{AUC}_{\text{localLaplace}}} \approx \sqrt{3}$. Our results as shown in Table ~\ref{tab:std_auc_w_accurate_P_and_N} (in the Appendix) verified this observation. 


\textbf{Sensitivity of $\alpha$.} We conduct experiments to check the sensitivity of $\alpha$ for \ourapplap{}. Given a total DP budget $\epsilon$, $\epsilon_{localSum} = \alpha \epsilon$ and  $\epsilon_{localP} = (1-\alpha \epsilon)$. We assign the data samples uniformly to $1,000$ clients (IID setting) and on average each client has 458 data samples. The corresponding results are shown in  Figure ~\ref{fig:auc_dataset_size_alpha_criteo} (b). It shows that when $\alpha \in [0.5, 0.6]$, the corresponding standard deviation of AUC estimation is the smallest. Hence, for the following experiments, we set $\alpha=0.5$. Other settings (i.e. Non-IID and different number of clients) share the same pattern. To save space, we put the corresponding results in the Appendix as shown in Figure ~\ref{fig:std_auc_vs_dp_allocations_non_iid_clients_1000_criteo} and ~\ref{fig:std_auc_vs_dp_allocations_non_iid_clients_1000_eps_4_criteo}.


\textbf{Advantages of Adaptive Allocation.} We conduct some experiments to show the advantages of orthogonalization and corresponding adaptive allocation. Given the total DP budget $\epsilon=1$, we assign $1,000$ Non-IID samples to each client. 
We compared another approach which only does orthogonalization (without adaptive allocation). We tune the corresponding allocation parameter $\beta$ and let each client share the same $\beta$. The corresponding result is shown in  Figure ~\ref{fig:auc_dataset_size_alpha_criteo} (c) with legend "Orthogonal". It can achieve a smaller standard deviation than the naive one (tuning $\alpha$ for \ourapplap{}). The legend "Adaptive" in in  Figure ~\ref{fig:auc_dataset_size_alpha_criteo} (c) represents that each client adaptive allocates the DP budget and hence it should be a point in the figure. To make a better comparison with "Naive" and "Orthogonal", we show a horizontal line in the figure for "Adaptive" instead. Adaptive allocation can achieve the smallest standard deviation of AUC, which demonstrates the effectiveness of adaptive allocation. Similar results on the IID setting can be seen in the Figure ~\ref{fig:naive_orthogonal_adaptive_group_size_1000_eps_1}. The distribution of the adaptive allocated parameter $\beta$ for all clients can be seen in Figure ~\ref{fig:dist_of_local_beta} in the Appendix.

\textbf{RR vs. GlobalLaplace vs. LocalLaplace (w/wo. adaptive allocation).}  We now compare our proposed methods (RR, GlobalLaplace, and LocalLaplace (w/wo adaptive allocation)) in Table \ref{tab:std_auc_w_noisy_P_and_N} with different settings (i.e. different number of clients and different privacy budget $\epsilon$). We set the DP budget allocation parameter $\alpha = 0.5$ for LocalLaplace without adaptive allocation. It shows that RR can gain some advantages when the avg. $\#$ data samples per client is small. However, with increasing avg. $\#$ data samples per client, LocalLaplace can achieve smaller standard deviation of AUC estimation since it adds smaller amounts of noise than RR and GlobalLaplace.  LocalLaplace with adaptive allocation can perform the best under most of the cases.

\begin{table}[!htp]\centering
\scriptsize
\setlength{\tabcolsep}{0.5em} 
{\renewcommand{\arraystretch}{1.2}
\begin{tabular}{c|c|c|c|c|c}\toprule
&$\#$ clients &10 &458 &4,584 &45,840 \\ \hline 
&avg. $\#$ data samples per client &45,840.00 &1,000.00 &100.00 &10.00 \\ \hline 
\multirow{6}{*}{$\epsilon=1.0$} &RR &\multicolumn{4}{c}{2.17e-3} \\ \cline{2-6}
&GlobalLaplace &1.22e-4 &8.48e-4 &2.39e-3 &8.08e-3 \\ \cline{2-6}
&LocalLaplace (IID) &1.13e-4 &9.64e-4 &2.26e-3 &7.39e-3 \\ \cline{2-6}
&LocalLaplace (Non-IID) &8.98e-5 &5.29e-4 &1.86e-3 &5.45e-3 \\ \cline{2-6}
&adaptive (IID) &5.15e-5 &3.92e-4 &1.22e-3 &3.80e-3 \\ \cline{2-6}
&LocalLaplace adaptive (Non-IID) &2.93e-5 &1.22e-4 &3.92e-4 &1.03e-3 \\ \hline 
\multirow{6}{*}{$\epsilon=2.0$} &RR &\multicolumn{4}{c}{1.02e-3} \\ \cline{2-6}
&GlobalLaplace &5.85e-5 &4.15e-4 &1.28e-3 &3.95e-3 \\ \cline{2-6}
&LocalLaplace (IID) &5.74e-5 &4.72e-4 &1.21e-3 &4.09e-3 \\ \cline{2-6}
&LocalLaplace (Non-IID) &4.59e-5 &2.91e-4 &9.45e-4 &3.09e-3 \\ \cline{2-6}
&LocalLaplace adaptive (IID) &2.60e-5 &1.84e-4 &6.48e-4 &1.66e-3 \\ \cline{2-6}
&LocalLaplace adaptive (Non-IID) &1.59e-5 &6.01e-5 &1.52e-4 &5.54e-4 \\ \hline 
\multirow{6}{*}{$\epsilon=4.0$} &RR &\multicolumn{4}{c}{3.49e-4} \\ \cline{2-6}
&GlobalLaplace &2.92e-5 &2.09e-4 &6.58e-4 &1.88e-3 \\ \cline{2-6}
&LocalLaplace (IID) &3.11e-5 &1.94e-4 &5.80e-4 &1.86e-3 \\ \cline{2-6}
&LocalLaplace (Non-IID) &2.49e-5 &1.48e-4 &4.44e-4 &1.53e-3 \\ \cline{2-6}
&LocalLaplace adaptive (IID) &1.36e-5 &8.20e-5 &2.87e-4 &9.08e-4 \\ \cline{2-6}
&LocalLaplace adaptive (Non-IID) &7.80e-6 &2.81e-5 &8.72e-5 &2.77e-4 \\ \hline 
\multirow{6}{*}{$\epsilon=8.0$} &RR &\multicolumn{4}{c}{4.41e-5} \\ \cline{2-6}
&GlobalLaplace &1.54e-5 &9.82e-5 &3.31e-4 &1.05e-3 \\ \cline{2-6}
&LocalLaplace (IID) &1.51e-5 &1.06e-4 &3.27e-4 &8.52e-4 \\ \cline{2-6}
&LocalLaplace (Non-IID) &1.06e-5 &6.99e-5 &2.19e-4 &7.50e-4 \\ \cline{2-6}
&LocalLaplace adaptive (IID) &7.01e-6 &4.55e-5 &1.37e-4 &4.61e-4 \\ \cline{2-6}
&LocalLaplace adaptive (Non-IID) &3.89e-6 &1.36e-5 &4.34e-5 &1.38e-4 \\
\bottomrule
\end{tabular}}
\caption{Standard deviation of AUC estimated by different methods (DP budget allocation $\alpha = 0.5$ which means that $\epsilon_{\text{localSum}} = \epsilon_{\text{localP}} = 0.5\epsilon$). IID: assigned samples to clients uniformly. Non-IID: assigned samples to clients based on their prediction scores.}\label{tab:std_auc_rr_global_local_adaptive}
\end{table}


\section{Related Work}
\label{sec:related_work}

\para{Federated Learning.} FL~\cite{mmr+17,yang2019federated} can be mainly classified into three categories: \textit{horizontal FL},  \textit{vFL}, and \textit{federated transfer learning} ~\cite{yang2019federated}. 
When the jointly trained model is a neural network, the setting is the same as split learning such as SplitNN ~\cite{vepakomma2018split}. 

\textbf{Information Leakage in vFL.} Recently, studies show that in vFL, even though the raw data (feature and label) is not shared,
sensitive information can still be leaked from the gradients and intermediate embeddings communicated between parties. For example, \cite{vepakomma2019reducing} and \cite{sunDefending2021} showed that server's raw features can be leaked from the forward cut layer embedding.  In addition, \cite{li2021label} studied the label leakage problem but the leakage source was the backward gradients rather than forward embeddings. 

\textbf{Information Protection in vFL.} 
There are three main categories of information protection techniques in vFL:
\textbf{1)} cryptographic methods such as {secure multi-party computation} \cite{bonawitz2017practical}; \textbf{2)} system-based methods including {trusted execution environments} \cite{subramanyan2017formal}; and \textbf{3)} perturbation methods that add noise to the communicated messages~\cite{abadi2016deep, mcmahan2017learning,erlingsson2019amplification,cheu2019distributed,zhu2019dlg}. In this paper, we focus on adding DP noise to protect the private label information during computing AUC.
\section{Conclusion}
\label{sec:conclusions}


We propose two label DP  algorithms to compute AUC when evaluating models in vFL protocol. Through experiments, we find our algorithms can estimate AUCs accurately.
We primarily focus on the online advertising scenario when the predicted scores are not considered as sensitive, and we leave the generalization to other scenarios for future work. We hope our work can bring more attention to the current vFL literature on how to protect privacy when evaluating models.



\clearpage
\bibliographystyle{plain}
\bibliography{references}



\clearpage

\appendix
\section{Appendix}
\label{sec:appendix}

\textbf{Appendix Outline:}

Section ~\ref{sec:illustration_rr}: Illustration of our proposed \ourapp{} with randomized response.

Section ~\ref{sec:illustration_laplace}: Illustration of our proposed \ourapp{} with Gaussian or Laplace mechanism

Section ~\ref{sec:algo_rr_appendix}: Algorithm  ~\ref{alg:clients_flipping_labels}: Clients leverages randomized response to flip labels with $\epsilon$-LabelDP guarantee

Section ~\ref{sec:algorithm_noisy_auc_appendix} : Algorithm ~\ref{alg:clients_server_cal_auc}: Sever computes noisy AUC with clients' local flipped labels

Section ~\ref{sec:clean_auc_appendix}: Algorithm ~\ref{alg:real_auc_from_corr}: Server computes the clean/final AUC

Section ~\ref{sec:algorithm_noisy_auc_lap_appendix}: Algorithm ~\ref{alg:clients_server_cal_auc_2}: Sever computes noisy AUC with clients' local labels

Section ~\ref{sec:auc_probabilistic_view}: Computing  AUC from a probabilistic perspective

Section ~\ref{sec:utility_rr}: Utility analysis of RR

Section ~\ref{sec:auc_probabilistic_view}:  Computing  AUC from a probabilistic perspective

Section ~\ref{sec:utility_rr}: Utility analysis of RR

Section ~\ref{sec:std_auc_different_alpha_non_iid_appendix}: Standard deviation of AUC with different DP budget allocation $\alpha$ in the Non-IID setting

Section ~\ref{sec:ortho_adaptive_iid_appendix}:  Orthogonalizatin vs. Adaptive allocation in the IID setting

Section ~\ref{sec:noisy_auc_rr_appendix}: Noisy AUC computed by Randomized Response without debiasing

Section ~\ref{sec:more_results_criteo_rr_appendix}: More results of Randomized Responses on Criteo dataset

Section ~\ref{sec:var_auc_globallaplace}: Variance of AUC estimated by GlobalLaplace

Section ~\ref{sec:utility_analysis_locallap_appendix}: Utility analysis of LocalLaplace

Section ~\ref{sec:estimate_pi}: Estimate $\pi'$

Section ~\ref{sec:compare_all_methods_auc_appendix}: Table ~\ref{tab:std_auc_w_noisy_P_and_N}: std of AUC estimated different methods: RR vs. GlobalLaplace vs. LocalLapace with adaptive budget allocation

Section ~\ref{sec:sorting_module}: Privacy analysis of prediction scores 

Section ~\ref{sec:dist_of_local_beta}: Figure: ~\ref{fig:dist_of_local_beta}: Distribution of local $\beta$

Section ~\ref{sec:computation_resources}: Computational resources

\clearpage

\section{Illustration of our proposed \ourapp{} with randomized response}
\label{sec:illustration_rr}

\begin{figure*}[ht!]
  \centering
  \includegraphics[width=1.0\linewidth]{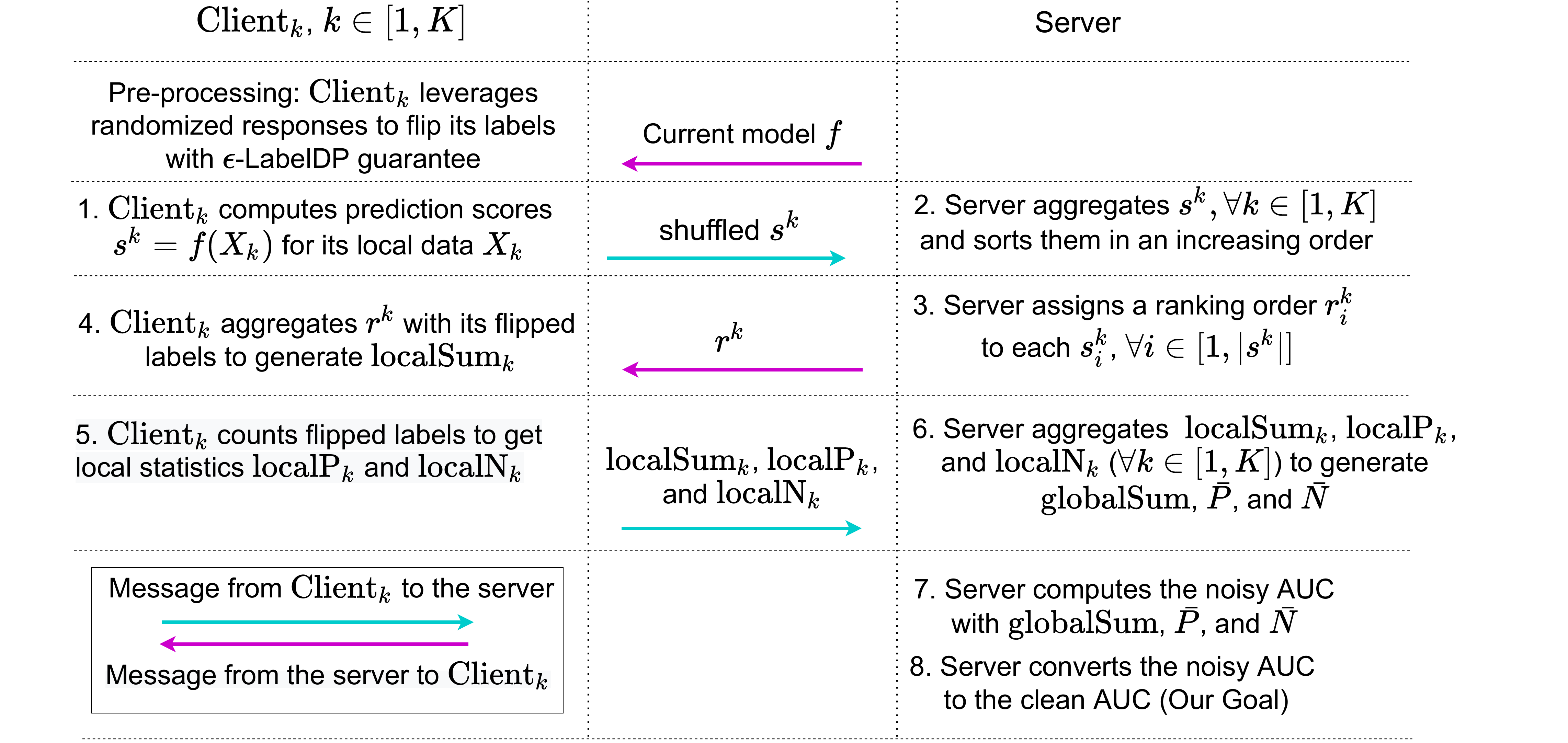}
  \caption{Illustration of our proposed \ourapp{} with randomized response.}
 \label{fig:fedauc_illustration_rr} 
 \end{figure*}

Illustration of our proposed \ourapp{} with randomized response can be seen in Figure ~\ref{fig:fedauc_illustration_rr}.

\section{Illustration of our proposed \ourapp{} with Gaussian or Laplace mechanism}
\label{sec:illustration_laplace} 

\begin{figure*}[ht!]
  \centering
  \includegraphics[width=1.0\linewidth]{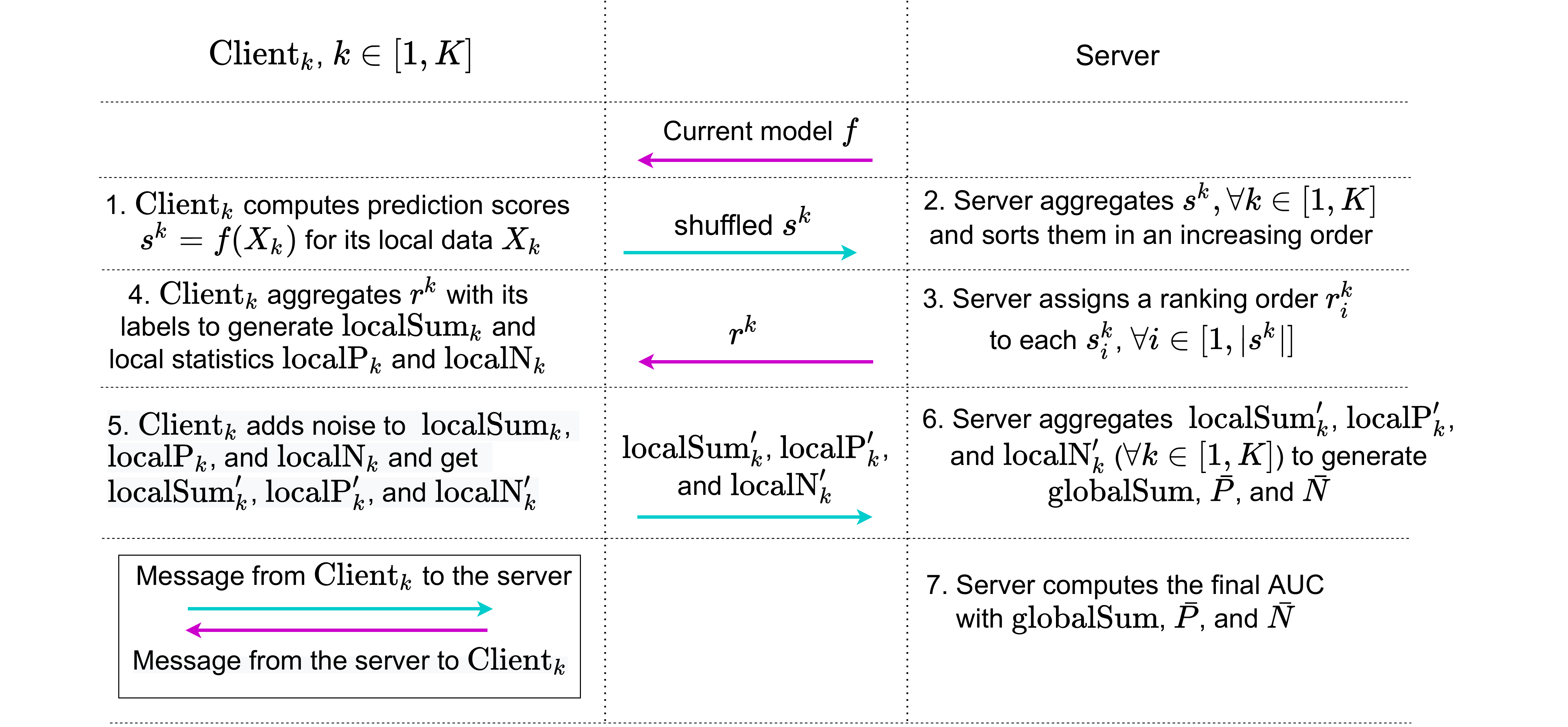}
  \caption{Illustration of our proposed \ourapp{} with Gaussian or Laplace mechanism.}
 \label{fig:fedauc_illustration_lap} 
 \end{figure*}

Illustration of our proposed \ourapp{} with Gaussian or Laplace mechanism can be seen in Figure ~\ref{fig:fedauc_illustration_lap}.

\section{Algorithm  ~\ref{alg:clients_flipping_labels}: Clients leverages randomized response to flip labels with $\epsilon$-LabelDP guarantee}
\label{sec:algo_rr_appendix}

Algorithm ~\ref{alg:clients_flipping_labels} shows how clients leverages randomized response to flip labels with $\epsilon$-LabelDP guarantee.

\begin{algorithm}[ht!]
\caption{Clients leverages randomized response to flip labels with $\epsilon$-LabelDP guarantee}\label{alg:clients_flipping_labels}
\KwData{The $K$ clients are index by $k$. Each client $C_k$ has data $D_k$ with $(X_k, Y_k)$ where $Y_k \in [0,1]$. Privacy budget $\epsilon$.}
\KwResult{Each client $C_k$ will have noisy data $D_k'$ with $\epsilon$-LabelDP guarantee}
 \For{each client $C_k$}{
    \For{each data point $(x_i, y_i) \in (X_k, Y_k)$}{
    replace $y_i$ by $\tilde{y_i}$ with randomized response as defined in equation ~\ref{eq:rr}.
    }
}
\end{algorithm}

\section{Algorithm ~\ref{alg:clients_server_cal_auc}: Sever Computes Noisy AUC with Clients' Local Flipped Labels}
\label{sec:algorithm_noisy_auc_appendix}

Algorithm ~\ref{alg:clients_server_cal_auc} shows how sever compute the noisy AUC with clients' local flipped labels.

\begin{algorithm}[ht!]
\caption{Sever Computes Corrupted AUC with Clients' Local Flipped Labels}\label{alg:clients_server_cal_auc}
\KwData{The $K$ clients are index by $k$. Each client $C_k$ has noisy data $D_k'$ with $(X_k, Y_k')$ where $Y_k' \in [0,1]$. Model $f$.}
\KwResult{Corrupted AUC: $\text{AUC}^{D_\text{noisy}}$}

\textbf{ // Clients Execute}

     \For{each client $C_k$}{
        \For{each data point $(x_i, y_i') \in (X_k, Y_k'
        )$}{
        Calculate the corresponding prediction score $s_i^k = f(x_i)$
        }
     Send all prediction scores $s^k$ to the server.
    }

\textbf{ // Server Executes}

Aggregate all the prediction scores and sort them in an increasing order.

Each prediction score $s_i^k$ ($i \in [1, M]$) will be assigned a ranking order $r_i^k$ as described in Section ~\ref{sec:computing_auc_ranking}.

Sends each ranking order $r_i^k$ back to the corresponding client $C_k$ which owns $s_i^k$.

\textbf{ // Clients Execute}
    
     \For{each client $C_k$}{
     
     $\text{localSum}_k, \text{localP}_k, \text{localN}_k  =  0, 0, 0 $

    \For{each data point $(x_i, y_i') \in (X_k, Y_k')$}{
        $\text{localSum}_k += r_i^k * y_i'$
        
        \eIf{$y_i' == 1$}{
            $\text{localP}_k  += 1$
                        }
            {
            $\text{localN}_k  += 1$
            }
        }
     $C_k$ sends $\text{localSum}_k$, $ \text{localP}_k$, and $\text{localN}_k$ to the server.
    }

\textbf{ // Server Executes}

// Server aggregates all the $\text{localSum}_k$, $ \text{localP}_k$, and $\text{localN}_k$.

$\text{globalSum} = \sum_{k}^K  \text{localSum}_k$

$\bar{P} = \sum_{k}^K \text{localP}_k$

$\bar{N} = \sum_{k}^K \text{localN}_k$

$\text{AUC}^{D_{\text{noisy}}} = ({\text{globalSum} - {\bar{P}(\bar{P}-1)}/{2}}) / ({\bar{P}\bar{N}}$)

\end{algorithm}

\section{Algorithm ~\ref{alg:real_auc_from_corr}: Server computes the clean/final AUC}
\label{sec:clean_auc_appendix}

Algorithm ~\ref{alg:real_auc_from_corr} shows how the server compute the clean/final AUC.

\begin{algorithm}[ht!]
\caption{Server computes the clean/final AUC}\label{alg:real_auc_from_corr}
\KwData{Corrupted AUC $\text{AUC}^{D_{\text{noisy}}}$, $\bar{P}$, $\bar{N}$, LabelDP budget $\epsilon$}
\KwResult{$\text{AUC}^{D_{\text{clean}}}$}

\textbf{ // Server Executes}

$\rho_{+} = \rho_{-} = \frac{1}{1+ e^{\epsilon}}$

$P' = \frac{\bar{P}(1-\rho_{-}) - \bar{N}\rho_{-}}{1-\rho_{+}-\rho_{-}}$

$N'= \bar{P}+\bar{N}-P'$

$\pi' = \frac{P'}{P' + N'}$

$\alpha = \frac{(1-\pi') \rho_{-}}{\pi(1-\rho_{+}) + (1-\pi')\rho_{-}}$

$\beta = \frac{\pi' \rho_{+}}{\pi' \rho_{+} + (1-\pi')(1-\rho_{-})}$
    
$ \text{AUC}^{D_{\text{clean}}}  =  \frac{\text{AUC}^{D_{\text{noisy}}} - \frac{\alpha+\beta}{2}}{1-\alpha-\beta}    $
\end{algorithm}

\section{Algorithm ~\ref{alg:clients_server_cal_auc_2}: Sever Computes Noisy AUC with Clients' Local Labels}
\label{sec:algorithm_noisy_auc_lap_appendix}

Algorithm ~\ref{alg:clients_server_cal_auc_2} shows how server compute noisy AUC with clients' local labels.

 \begin{algorithm}[ht!]
\caption{Sever Computes Noisy AUC with Clients' Local  Labels}\label{alg:clients_server_cal_auc_2}
\KwData{The $K$ clients are index by $k$. Each client $C_k$ has  data $D_k$ with $(X_k, Y_k)$ where $Y_k \in [0,1]$. Model $f$.}
\KwResult{AUC: $\text{AUC}$}

\textbf{ // Clients Execute}

     \For{each client $C_k$}{
        \For{each data point $(x_i, y_i) \in (X_k, Y_k
        )$}{
        Calculate the corresponding prediction score $s_i^k = f(x_i)$
        }
     Send all prediction scores $s^k$ to the server.
    }

\textbf{ // Server Executes}

Aggregate all the prediction scores and sort them in an increasing order.

Each prediction score $s_i^k$ ($i \in [1, M]$) will be assigned a ranking order $r_i^k$ as described in Section ~\ref{sec:computing_auc_ranking}.

Sends each ranking order $r_i^k$ back to the corresponding client $C_k$ which owns $s_i^k$.

\textbf{ // Clients Execute}
    
     \For{each client $C_k$}{
     
     $\text{localSum}_k, \text{localP}_k, \text{localN}_k  =  0, 0, 0 $

    \For{each data point $(x_i, y_i') \in (X_k, Y_k')$}{
        $\text{localSum}_k += r_i^k * y_i$
        
        \eIf{$y_i == 1$}{
            $\text{localP}_k  += 1$
                        }
            {
            $\text{localN}_k  += 1$
            }
        }
        
      $\text{localSum}_k'$ = $\text{localSum}_k$ + noise
      
      $\text{localP}_k'$ = $ \text{localP}_k$ + noise
      
      $\text{localN}_k$ = $|Y_k| - \text{localP}_k'$ 
      
     $C_k$ sends $\text{localSum}_k'$, $ \text{localP}_k'$, and $\text{localN}_k'$ to the server.
    }

\textbf{ // Server Executes}

// Server aggregates all the $\text{localSum}_k'$, $ \text{localP}_k'$, and $\text{localN}_k'$.

$\text{globalSum} = \sum_{k}^K  \text{localSum}_k'$

$\bar{P} = \sum_{k}^K \text{localP}_k'$

$\bar{N} = \sum_{k}^K \text{localN}_k'$

$\text{AUC} = ({\text{globalSum} - {\bar{P}(\bar{P}-1)}/{2}}) / ({\bar{P}\bar{N}}$)

\end{algorithm}

\section{Computing  AUC from a probabilistic perspective}
\label{sec:auc_probabilistic_view}

In a binary classification problem, given a threshold $\theta$, a predicted score $s_i$ is predicted to be $1$ if $s_i \ge \theta$. Given the ground-truth label and the predicted label (at a given threshold $\theta$), we can quantify the accuracy of the classifier on the dataset with True positives (TP($\theta$)), False positives (FP($\theta$)), False negatives (FN($\theta$)), and True negatives (TN($\theta$)). Area under the Receiver operating characteristic (ROC) curves plots two variables: True Positive Rate (TPR)
 False Positive Rate (FPR). TPR (i.e. recall) is defined as $TPR(\theta) = \frac{TP(\theta)}{TP(\theta) + FN(\theta)}$. False Positive Rate (FPR) is defined as $(\theta) = \frac{FP(\theta)}{FP(\theta) + TN(\theta)}$.

\begin{itemize}
    \item True positives, TP($\theta$), are the data points in test whose true label and predicted label equals $1$. \textit{i.e.} $y_i = 1$ and $s_i \ge \theta$
    \item False positives, FP($\theta$), are the data points in test whose true label is $0$ but the predicted label is $1$. \textit{i.e.} $y_i = 0$ and $s_i \ge \theta$.
  \item  False negatives, FN($\theta$), are data points whose true label is $1$ but the predicted label is $0$. \textit{i.e.} $y_i = 1$ and $s_i < \theta$. 
  \item True negatives, TN($\theta$), are data points whose true label is $0$ and the predicted label is $0$. \textit{i.e.} $y_i = 0$ and $s_i < \theta$. 
\end{itemize}


The ROC curve is defined by plotting  pairs of FPR($\theta$) (x-axis) versus TPR($\theta$) (y-axis) over all possible thresholds $\theta$. ROC curve starts at $(0,0)$ and ends at $(1,1)$. The area under the ROC curve (AUC) is used to evaluate the performance of a binary classifier. If the classifier is good, the ROC curve will be close to the left and upper boundary and AUC will be close to $1.0$. On the other hand, if the classifier is poor, the ROC curve will be close to  line from $(0,0)$ to $(1,1)$ with AUC around $0.5$.

\subsection{Computing AUC with Time Complexity O(PN)}

Area under ROC curve (AUC) is well studied in statistics and is equivalent to Mann-Whitney U statistics ~\cite{AUCProb2002,dpforclassifierEvaluation}. We focus on computing the AUC by 
 viewing it as the probability of correct ranking of a random positive-negative pair. Suppose we have $M$ samples including $P$ positives and $N$ negatives, where $M = P + N$. Given a classifier $\mathcal{F}$, it gives a prediction score $s_i$ for each sample $i \in [1, M]$. We then have $MN$ positive-negative pairs with score $<s_i, s_j>$ ($i \in [1, M]$ and $j \in [1, N]$) where $s_i$ and $s_j$ are prediction scores of positive sample $i$ and negative sample $j$ respectively. Then AUC of classifier $\mathcal{F}$ is:

\begin{equation}
    \text{AUC} = \frac{I(<s_i, s_j>)}{MN}, i \in [1, M] \quad \text{and} \quad j \in [1, N]
\end{equation}

where 
\begin{equation*}
    I(<s_i, s_j>) = \begin{cases}
                    1,  \text{if} \quad s_i > s_j,\\
                    0, otherwise.
                    \end{cases}
\end{equation*}

The time complexity of computing AUC is $O(PN)$.

\subsection{Computing AUC with Time Complexity $O(M \log M)$}
\label{sec:computing_auc_ranking}

We firstly rank all instances based on their prediction scores in an increasing order. Each score $s_i$ ($i \in [1, M]$) will be assigned a ranking $r_i$. The instance with the highest order will be assigned $r = M - 1$, and the second highest one will be assigned $M -2$ and so on. The smallest ranking score is $r = 0$.

Then AUC can be computed as:

\begin{equation}
\label{eq:auc_ranking_appendix}
    \text{AUC} = \frac{\sum_{i=1}^{M} r_i \cdot y_i - \frac{P(P-1)}{2} }{PN}
\end{equation}

where $y_i \in \{0, 1\}$ is the ground-truth label for instance $i$ (with ranking score $r_i$). The time complexity of this method is $O(M\log M)$.

\section{Utility analysis of RR}
\label{sec:utility_rr}



We analyze the  variance of the computed AUC. To simplify the analysis, we use the accurate $P$ and $N$ to compute the AUC \footnote{Empirically, we did not find too much  difference between using estimated and accurate $P$ and $N$}. Let the flipping probability as $r = \frac{1}{1+exp(\epsilon)}$, then we have



\begin{equation}
\label{eq:std_of_noisy_auc_w_accurate_p_n}
\begin{split}
\textbf{Var}(AUC^{D_{\text{clean}}})  \\ = \textbf{Var}(\frac{\sum^K sum_i - \frac{P * (P-1)}{2}}{P * N})  =  \textbf{Var}(\frac{\sum^K sum_i}{P * N}) \\= \textbf{Var}(\frac{\sum_{i=1}^P s_i (1-r)+\sum_{j=1}^N s_j r}{P * N}) \\= \frac{\sum_{i=1}^{M}s_i^2 r  (1-r)}{P^2 * N^2} = \frac{r(1-r)\sum_{i=1}^{M-1}i^2}{P^2 * N^2} \\
= r(1-r)\frac{M(M-1)(2M-1)/6}{P^2 * N^2} \\
= \frac{exp(\epsilon)}{(1+exp(\epsilon))^2}\frac{M(M-1)(2M-1)/6}{P^2 * N^2}
\end{split}
\end{equation}

\begin{table*}[ht!]\centering
\setlength{\tabcolsep}{0.5em} 
{\renewcommand{\arraystretch}{1.2}
\begin{tabular}{c|c|c|c|c|c|c}\toprule
& & avg. $\#$ data samples per client &45,840 &458.4 &4.6 &1 \\ \hline
\multirow{3}{*}{$\epsilon=1.0$} &GlobalLaplace &- &4.84e-5 &5.26e-4 &4.45e-3 &1.10e-2 \\ \cline{2-7}
&\multirow{2}{*}{LocalLaplace} &i.i.d &4.81e-5 &4.66e-4 &4.25e-3 &6.44e-3 \\
& &\cellcolor[HTML]{A8A8A8}non i.i.d &\cellcolor[HTML]{A8A8A8}3.40e-5 &\cellcolor[HTML]{A8A8A8}2.98e-4 &\cellcolor[HTML]{A8A8A8}2.76e-3 &\cellcolor[HTML]{A8A8A8}6.08e-3 \\  \hline
\multirow{3}{*}{$\epsilon=2.0$} &GlobalLaplace &- &2.61e-5 &2.58e-4 &2.37e-3 &5.65e-3 \\ \cline{2-7}
&\multirow{2}{*}{LocalLaplace} &i.i.d &2.61e-5 &2.48e-4 &2.10e-3 &2.87e-3 \\
& &\cellcolor[HTML]{A8A8A8}non i.i.d &\cellcolor[HTML]{A8A8A8}1.48e-5 &\cellcolor[HTML]{A8A8A8}1.51e-4 &\cellcolor[HTML]{A8A8A8}1.38e-3 &\cellcolor[HTML]{A8A8A8}3.20e-3 \\  \hline
\multirow{3}{*}{$\epsilon=4.0$} &GlobalLaplace &- &1.30e-5 &1.21e-4 &1.21e-3 &2.68e-3 \\ \cline{2-7}
&\multirow{2}{*}{LocalLaplace} &i.i.d &1.23e-5 &1.20e-4 &1.14e-3 &1.56e-3 \\
& &\cellcolor[HTML]{A8A8A8}non i.i.d &\cellcolor[HTML]{A8A8A8}6.63e-6 &\cellcolor[HTML]{A8A8A8}7.22e-5 &\cellcolor[HTML]{A8A8A8}6.20e-4 &\cellcolor[HTML]{A8A8A8}1.69e-3 \\  \hline
\multirow{3}{*}{$\epsilon=8.0$} &GlobalLaplace &- &6.29e-6 &5.94e-5 &6.12e-4 &1.34e-3 \\ \cline{2-7}
&\multirow{2}{*}{LocalLaplace} &i.i.d &5.79e-6 &6.32e-5 &5.07e-4 &8.09e-4 \\
& &\cellcolor[HTML]{A8A8A8}non i.i.d &\cellcolor[HTML]{A8A8A8}3.98e-6 &\cellcolor[HTML]{A8A8A8}3.70e-5 &\cellcolor[HTML]{A8A8A8}3.44e-4 &\cellcolor[HTML]{A8A8A8}8.53e-4 \\
\bottomrule
\end{tabular}}
\caption{Standard deviation of estimated AUC with accurate $P$ and $N$. i.i.d: assigned samples to clients uniformly. non i.i.d: assigned samples to clients based on their prediction scores.}\label{tab:std_auc_w_accurate_P_and_N}
\end{table*}

 \section{Standard deviation of AUC with different DP budget allocation $\alpha$ in the non iid setting}
 \label{sec:std_auc_different_alpha_non_iid_appendix}

\begin{figure*}[ht!]
  \centering
  \includegraphics[width=0.5\linewidth]{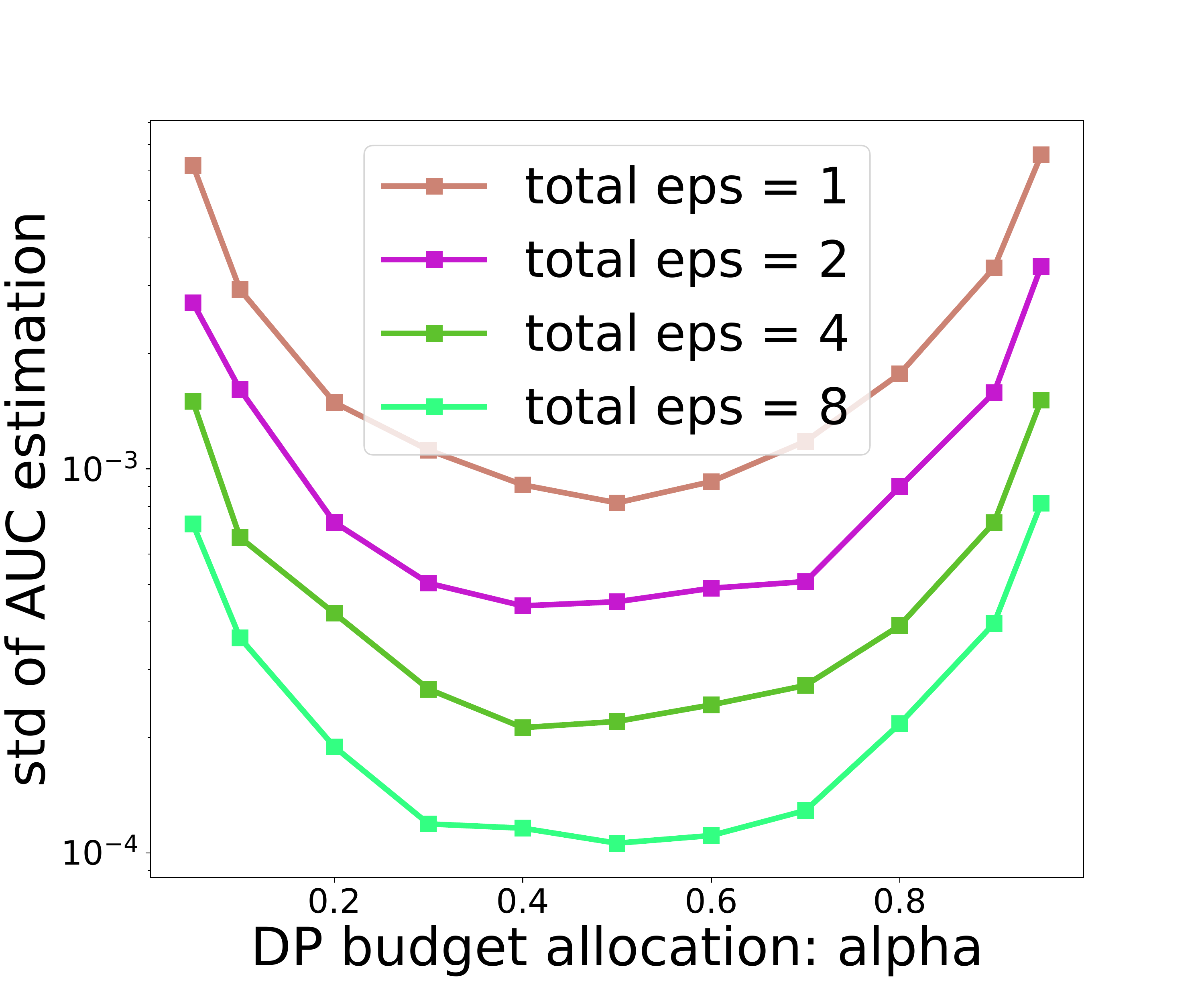}
  \caption{Standard deviation of AUC estimation with different allocation $\alpha$ under the Non-IID setting.}
 \label{fig:std_auc_vs_dp_allocations_non_iid_clients_1000_criteo} 
 \end{figure*}

Figure ~\ref{fig:std_auc_vs_dp_allocations_non_iid_clients_1000_criteo} shows the standard deviation of AUC estimation with different allocation $\alpha$ under the Non-IID setting (1000 clients).

\begin{figure*}[ht!]
  \centering
  \includegraphics[width=0.5\linewidth]{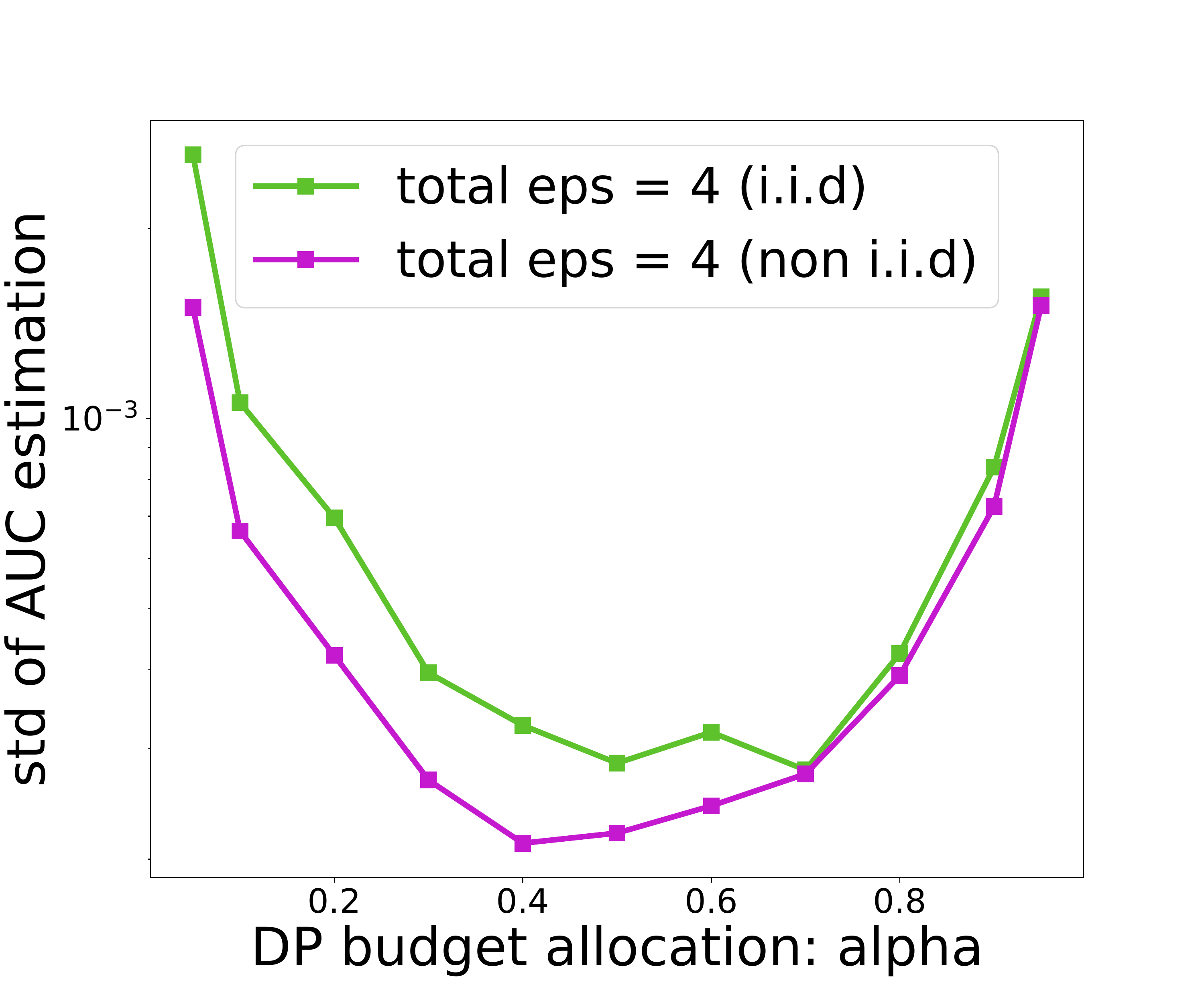}
  \caption{Standard deviation of AUC estimation with different allocation $\alpha$ under the Non-IID and IID setting with $\epsilon=4.0$.}
 \label{fig:std_auc_vs_dp_allocations_non_iid_clients_1000_eps_4_criteo} 
 \end{figure*}

Figure ~\ref{fig:std_auc_vs_dp_allocations_non_iid_clients_1000_eps_4_criteo}: Standard deviation of AUC estimation with different allocation $\alpha$ under the Non-IID and IID setting with $\epsilon=4.0$ (1000 clients).
 
 \section{Orthogonalizatin vs. Adaptive allocation in the IID setting}
 \label{sec:ortho_adaptive_iid_appendix}
 
\begin{figure*}[ht!]
  \centering
  \includegraphics[width=0.5\linewidth]{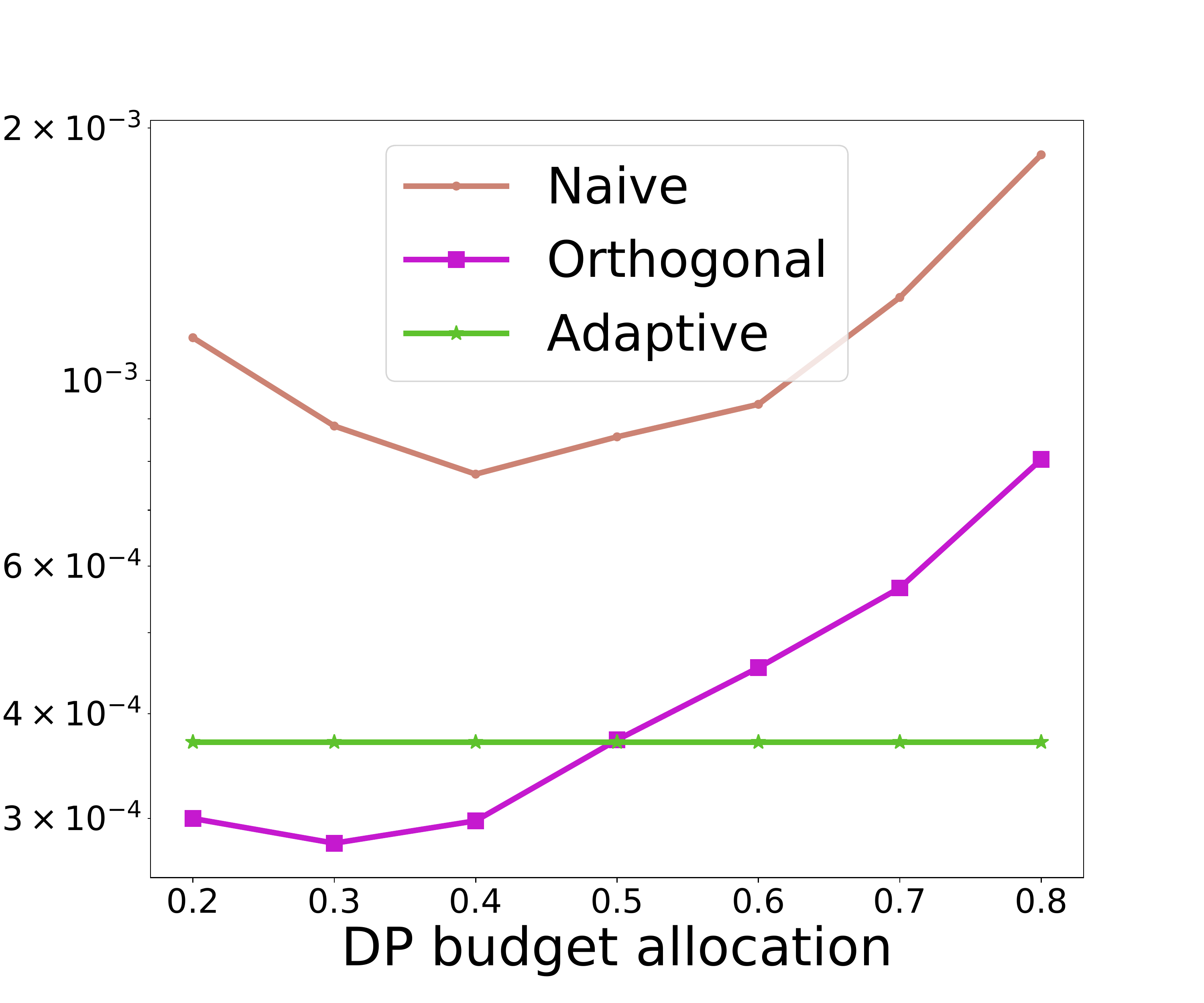}
  \caption{std of AUC estimation with different allocation methods (Naive vs. Orthogonal vs. Adaptive) in the IID setting (1000 sampler per client) with total privacy budget $\epsilon=1$.}
 \label{fig:naive_orthogonal_adaptive_group_size_1000_eps_1} 
 \end{figure*}

 Figure ~\ref{fig:naive_orthogonal_adaptive_group_size_1000_eps_1} shows the std of AUC estimation with different allocation methods (Naive vs. Orthogonal vs. Adaptive) in the IID setting (1000 sampler per client) with total privacy budget $\epsilon=1$.
 
\section{Noisy AUC computed by Randomized Response without debiasing}
\label{sec:noisy_auc_rr_appendix}

Table ~\ref{tab:noisy_auc_randomized_response_wo_converting} shows the noisy AUC computed by randomized response only without converting it to the clean AUC (without debiasing). The utility dropping a lot with small privacy budget $\epsilon$.

\begin{table}[ht!]\centering
\setlength{\tabcolsep}{0.5em} 
{\renewcommand{\arraystretch}{1.2}
\begin{tabular}{c|c|c|c|c|c}\toprule
AUC &$\epsilon$ &$\rho_{+}=\rho_{-}$ &Epoch 1 &Epoch 3 &Epoch 3 \\ \hline 
\multirow{6}{*}{Noisy AUC} &0.1 &47.50\% &0.509429 &0.510137 &0.510237 \\  \cline{2-6}
&0.5 &37.75\% &0.546986 &0.550344 &0.550862 \\ \cline{2-6}
&1 &26.89\% &0.592421 &0.598951 &0.600010 \\ \cline{2-6}
&2 &11.92\% &0.667666 &0.679436 &0.681911 \\ \cline{2-6}
&4 &1.80\% &0.735230 &0.751306 &0.754892 \\ \cline{2-6}
&10 &4.54e-5 &0.749343 &0.766438 &0.770177 \\ \hline 
scikit-learn &- &0 &0.749383 &0.766477 &0.770219 \\ \hline 
Tensorflow &- &0 &0.749382 &0.766478 &0.770219 \\
\bottomrule
\end{tabular}}
\caption{Noisy AUC computed by Randomized Response mechanism without debiasing.}\label{tab:noisy_auc_randomized_response_wo_converting}
\scriptsize
\end{table}

 \section{More results of Randomized Responses on Criteo dataset}
 \label{sec:more_results_criteo_rr_appendix}

The statistics of Criteo dataset can be seen Table ~\ref{tab:evaluation_set_statistics}. The corresponding results on small and large evaluation set can be seen in Table ~\ref{tab:auc_std_criteo} and ~\ref{tab:auc_std_criteo_large} respectively. 

\begin{table*}[ht!]\centering
\setlength{\tabcolsep}{0.5em} 
{\renewcommand{\arraystretch}{1.2}
\begin{tabular}{|c|c|c|c|c|}\toprule
& $\#$ Positives ($P$) & $\#$ Negatives ($N$) & $\#$ Total ($M$) \\ \hline
Small Criteo Evaluation Set &117,317 &341,090 &458,407 \\ \hline
Large Criteo Evaluation Set &1,173,981 &3,410,081 &4,584,062 \\ 
\bottomrule
\end{tabular}}
\caption{Statistics of Two Criteo Evaluation Sets}\label{tab:evaluation_set_statistics}
\end{table*}
 
\begin{table*}[ht!]\centering
\setlength{\tabcolsep}{0.5em} 
{\renewcommand{\arraystretch}{1.2}
\begin{tabular}{|c|c|c|c|c|c|c}\toprule
AUC and std &$\epsilon$ &$\rho_{+}=\rho_{-}$ &Epoch 1 &Epoch 2 &Epoch 3 \\ \hline
\multirow{10}{*}{\ourapprr{}} &0.1 &47.50\% &0.745667 $\pm$ 2.41$\text{e-}$2 &0.765312 $\pm$ 2.40$\text{e-}$2 &0.777627 $\pm$ 2.62$\text{e-}$2 \\ \cline{2-6}
&0.25 &43.78\% &0.749169 $\pm$ 8.69$\text{e-}$3 &0.767502 $\pm$ 8.76$\text{e-}$3 &0.771198 $\pm$ 7.57$\text{e-}$3 \\ \cline{2-6}
&0.5 &37.75\% &0.749206 $\pm$ 5.39$\text{e-}$3 &0.767081 $\pm$ 4.79$\text{e-}$3 &0.769621 $\pm$ 4.56$\text{e-}$3 \\ \cline{2-6}
&1 &26.89\% &0.749495 $\pm$ 2.11$\text{e-}$3 &0.766833 $\pm$ 2.09$\text{e-}$3 &0.770707 $\pm$ 2.23$\text{e-}$3 \\ \cline{2-6}
&2 &11.92\% &0.749453 $\pm$ 9.43$\text{e-}$4 &0.766621 $\pm$ 8.76$\text{e-}$4 &0.770548 $\pm$ 1.01$\text{e-}$3 \\ \cline{2-6}
&3 &4.74\% &0.749295 $\pm$ 5.46$\text{e-}$4 &0.766490 $\pm$ 4.62$\text{e-}$4 &0.770183 $\pm$ 6.14$\text{e-}$4 \\ \cline{2-6}
&4 &1.80\% &0.749207 $\pm$ 3.75$\text{e-}$4 &0.766458 $\pm$ 3.33$\text{e-}$4 &0.770161 $\pm$ 3.21$\text{e-}$4 \\ \cline{2-6}
&5 &0.67\% &0.749399 $\pm$ 2.05$\text{e-}$4 &0.766477 $\pm$ 2.14$\text{e-}$4 &0.770214 $\pm$ 1.33$\text{e-}$4 \\ \cline{2-6}
&10 &4.54$\text{e-}$5 &0.749383 $\pm$ 1.77$\text{e-}$5 &0.766475 $\pm$ 1.84$\text{e-}$5 &0.770214 $\pm$ 1.67$\text{e-}$5 \\ \cline{2-6}
&$+ \infty$ &0 &0.749383 $\pm$ 0.00$\text{e+}$0 &0.766477 $\pm$ 0.00$\text{e+}$0 &0.770219 $\pm$ 0.00$\text{e+}$0 \\ \hline
scikit-learn &- &0 &0.749383  $\pm$ 1.11$\text{e-}$16  &0.766477 $\pm$ 1.11$\text{e-}$16 &0.770219 $\pm$ 1.11$\text{e-}$16 \\ \hline
Tensorflow &- &0 &0.749382 $\pm$ 5.96$\text{e-}$8 &0.766478 $\pm$ 5.96$\text{e-}$8 &0.770219 $\pm$ 5.96$\text{e-}$8 \\
\hline
\end{tabular}} 
\caption{Mean and standard deviation (std) of AUC computed by \ourapprr{} and ground-truth on the Criteo evaluation set. }
\label{tab:auc_std_criteo}
\end{table*}

\begin{table*}[ht!]\centering
\setlength{\tabcolsep}{0.5em} 
{\renewcommand{\arraystretch}{1.2}
\begin{tabular}{|c|c|c|c|c|c|c|}\toprule
&$\epsilon$ &$\rho_{+}=\rho_{-}$ &Epoch 1 &Epoch 2 &Epoch 3 \\ \hline
\multirow{10}{*}{\ourapprr{}} &0.1 &47.50\% &0.751439\ $\pm$ 7.52$\text{e-}$3 &0.768640\ $\pm$ 6.96$\text{e-}$3 &0.774664\ $\pm$ 6.48$\text{e-}$3 \\ \cline{2-6}
&0.25 &43.78\% &0.750405\ $\pm$ 2.90$\text{e-}$3 &0.767958\ $\pm$ 2.85$\text{e-}$3 &0.771694\ $\pm$ 3.26$\text{e-}$3 \\ \cline{2-6}
&0.5 &37.75\% &0.750589\ $\pm$ 1.32$\text{e-}$3 &0.768324\ $\pm$ 1.62$\text{e-}$3 &0.772143\ $\pm$ 1.39$\text{e-}$3 \\ \cline{2-6}
&1 &26.89\% &0.750345\ $\pm$ 5.51$\text{e-}$4 &0.767988\ $\pm$ 7.27$\text{e-}$4 &0.771974\ $\pm$ 8.11$\text{e-}$4 \\ \cline{2-6}
&2 &11.92\% &0.750482\ $\pm$ 3.03$\text{e-}$4 &0.768001\ $\pm$ 3.03$\text{e-}$4 &0.772005\ $\pm$ 3.60$\text{e-}$4 \\ \cline{2-6}
&3 &4.74\% &0.750480\ $\pm$ 1.53$\text{e-}$4 &0.768043\ $\pm$ 1.82$\text{e-}$4 &0.772040\ $\pm$ 1.73$\text{e-}$4 \\ \cline{2-6}
&4 &1.80\% &0.750509\ $\pm$ 9.45$\text{e-}$5 &0.768055\ $\pm$ 1.11$\text{e-}$4 &0.772008\ $\pm$ 1.20$\text{e-}$4 \\ \cline{2-6}
&5 &0.67\% &0.750500\ $\pm$ 5.87$\text{e-}$5 &0.768041\ $\pm$ 6.73$\text{e-}$5 &0.772004\ $\pm$ 6.11$\text{e-}$5 \\ \cline{2-6}
&10 &4.54e-5 &0.750509\ $\pm$ 5.37$\text{e-}$6 &0.768046\ $\pm$ 5.83$\text{e-}$6 &0.772008\ $\pm$ 5.04$\text{e-}$6 \\ \cline{2-6}
&$+\infty$ &0 &0.750509\ $\pm$ 0.00$\text{e+}$0 &0.768045\ $\pm$ 1.11$\text{e-}$16 &0.772007\ $\pm$ 1.11$\text{e-}$16 \\ \hline
scikit-learn &- &0 &0.750509\ $\pm$ 1.11$\text{e-}$16 &0.768045\ $\pm$ 1.11$\text{e-}$16 &0.772007\ $\pm$ 1.11$\text{e-}$16 \\ \hline
Tensorflow &- &0 &0.750509\ $\pm$ 5.96$\text{e-}$8 &0.768044\ $\pm$ 5.96$\text{e-}$8 &0.772006\ $\pm$ 5.96$\text{e-}$8 \\
\bottomrule
\end{tabular}}
\caption{Mean and standard deviation (std) of AUC computed by FedAUC and baselines on large Criteo evaluation set.}\label{tab:auc_std_criteo_large}
\end{table*}

\section{Variance of AUC estimated by GlobalLaplace}
\label{sec:var_auc_globallaplace}

We also add a comparison partner \globallap{} here. All clients use $M - 1$ as the sensitivity when adding noise to their local statistics $\text{localSum}$. Given a privacy budget $\epsilon_{localSum}$, client $C_k$ draws the random noise from $Lap((M-1)/\epsilon_{localSum})$. Each client will add the same amount of noise to its $\text{localSum}$. Given $P$ and $N$ are accurate, the corresponding standard deviation of computed AUC is $std(AUC_{\text{\globallap{}}}) = \frac{\sqrt{2K}(M-1)}{P* N * \epsilon}$. Since
 
 \begin{equation}
 \label{eq:var_globallap}
 \textbf{Var}(AUC_{\text{\globallap{}}})   = \textbf{Var}(\frac{\sum^K sum_i - \frac{P * (P-1)}{2}}{P * N})  =  \textbf{Var}(\frac{\sum^K sum_i}{P * N}) = \frac{K * \sigma^2}{P^2 * N^2} = \frac{2K(M-1)^2}{P^2 * N^2 * \epsilon^2}
 \end{equation}

 \section{Utility analysis of LocalLaplace}
\label{sec:utility_analysis_locallap_appendix}

To simplify the analysis, we assume that $P$ and $N$ are accurate\footnote{only adding noise to $\text{localSum}$} and each client only has one data sample\footnote{$\text{sum}_i \in [1, K-1]$}, the standard deviation of computed AUC by \locallap{} is:

\begin{equation}
\label{eq:var_locallap}
   \textbf{Var} (\text{AUC}_{\text{\locallap{}}})  =  \textbf{Var}(\frac{\sum_{i=1}^K sum_i}{P  N}) =  \frac{\sum_{i=1}^{K-1} 2*i^2 / \epsilon^2}{P^2  N^2} = \frac{K(K-1)(2K-1)}{3P^2  N^2  \epsilon^2}
\end{equation} 
 
 \section{Estimate $\pi'$}
 \label{sec:estimate_pi}
 
 Suppose we observe $\bar{M}$ positive examples and $\bar{N}$ negative examples in the corrupted data $D'$. We have:

\begin{equation}
    P'+N' = \bar{P}+\bar{N}, \quad P'(1-\rho_{+}) + N \rho_{-} = \bar{P}
\end{equation}

where $P'$ and $N'$ are estimated positive and negative numbers in the clean data $D$.

We then obtain:

\begin{equation}
    P' = \frac{\bar{P}(1-\rho_{-}) - \bar{N}\rho_{-}}{1-\rho_{+}-\rho_{-}}, \quad N'= \bar{P}+\bar{N}-P'
\end{equation}

Then we use $P'$ and $N'$ to  estimate  base rate $\pi$ as $\pi'$:

\begin{equation}
    \pi' = \frac{P'}{P' + N'}
\end{equation}

\section{Table ~\ref{tab:std_auc_w_noisy_P_and_N}: std of AUC estimated different methods: RR vs. GlobalLaplace vs. LocalLapace with Adaptive Budget Allocation}
\label{sec:compare_all_methods_auc_appendix}

Table ~\ref{tab:std_auc_w_noisy_P_and_N} shows the std of AUC estimated by different methods (RR vs. GlobalLaplace vs. LocalLapace with Adaptive Budget Allocation).

\begin{table}[!htp]\centering
\scriptsize 
\setlength{\tabcolsep}{0.5em} 
{\renewcommand{\arraystretch}{1.2}
\begin{tabular}{c|c|c|c|c|c|c|c|c}\toprule
&$\#$ clients &10 &458 &1,000 &4,584 &45,840 &100,000 &458,407 \\ \hline 
&avg. $\#$ data samples per client &45,840.00 &1,000.00 &458.40 &100.00 &10.00 &4.60 &1 \\ \hline 
\multirow{4}{*}{$\epsilon=1.0$} &RR &\multicolumn{7}{c}{2.17e-3} \\ \cline{2-9}
&GlobalLaplace &1.22e-4 &8.48e-4 &1.24e-3 &2.39e-3 &8.08e-3 &1.24e-2 &2.77e-2 \\ \cline{2-9}
&LocalLaplace (IID) &1.13e-4 &9.64e-4 &1.26e-3 &2.26e-3 &7.39e-3 &1.10e-2 &1.99e-2 \\ \cline{2-9}
&\cellcolor[HTML]{A8A8A8}LocalLaplace (Non-IID) &\cellcolor[HTML]{A8A8A8}8.98e-5 &\cellcolor[HTML]{A8A8A8}5.29e-4 &\cellcolor[HTML]{A8A8A8}8.49e-4 &\cellcolor[HTML]{A8A8A8}1.86e-3 &\cellcolor[HTML]{A8A8A8}5.45e-3 &\cellcolor[HTML]{A8A8A8}8.33e-3 &\cellcolor[HTML]{A8A8A8}1.81e-2 \\ 
&adaptive (IID) &5.15e-5 &3.92e-4 & & 1.22e-3& 3.80e-3& & \\ \cline{2-9}
&adaptive (non-IID) &2.93e-5 &1.22e-4 & & 3.92e-4& 1.03e-3& & \\ \cline{2-9}
\hline 
\multirow{4}{*}{$\epsilon=2.0$} &RR &\multicolumn{7}{c}{1.02e-3} \\ \cline{2-9} 
&GlobalLaplace &5.85e-5 &4.15e-4 &5.62e-4 &1.28e-3 &3.95e-3 &5.21e-3 &1.32e-2 \\ \cline{2-9} 
&LocalLaplace (IID) &5.74e-5 &4.72e-4 &5.78e-4 &1.21e-3 &4.09e-3 &5.24e-3 &8.86e-3 \\ \cline{2-9}
&\cellcolor[HTML]{A8A8A8}LocalLaplace (Non-IID) &\cellcolor[HTML]{A8A8A8}4.59e-5 &\cellcolor[HTML]{A8A8A8}2.91e-4 &\cellcolor[HTML]{A8A8A8}4.93e-4 &\cellcolor[HTML]{A8A8A8}9.45e-4 &\cellcolor[HTML]{A8A8A8}3.09e-3 &\cellcolor[HTML]{A8A8A8}4.38e-3 &\cellcolor[HTML]{A8A8A8}9.96e-3 \\ 
&adaptive (IID) & 2.60e-5&1.84e-4 & & 6.48e-4& 1.66e-3& & \\ \cline{2-9}
&adaptive (non-IID) & 1.59e-5&6.01e-5 & & 1.52e-4& 5.54e-4& & \\ \cline{2-9}
\hline 
\multirow{4}{*}{$\epsilon=4.0$} &RR &\multicolumn{7}{c}{3.49e-4} \\ \cline{2-9} 
&GlobalLaplace &2.92e-5 &2.09e-4 &3.26e-4 &6.58e-4 &1.88e-3 &3.22e-3 &7.02e-3 \\ \cline{2-9} 
&LocalLaplace (IID) &3.11e-5 &1.94e-4 &3.00e-4 &5.80e-4 &1.86e-3 &2.40e-3 &4.38e-3 \\ \cline{2-9}
&\cellcolor[HTML]{A8A8A8}LocalLaplace (Non-IID) &\cellcolor[HTML]{A8A8A8}2.49e-5 &\cellcolor[HTML]{A8A8A8}1.48e-4 &\cellcolor[HTML]{A8A8A8}2.05e-4 &\cellcolor[HTML]{A8A8A8}4.44e-4 &\cellcolor[HTML]{A8A8A8}1.53e-3 &\cellcolor[HTML]{A8A8A8}2.26e-3 &\cellcolor[HTML]{A8A8A8}4.39e-3 \\ 
&adaptive (IID) & 1.36e-5&8.20e-5 & & 2.87e-4& 9.08e-4& & \\ \cline{2-9}
&adaptive (non-IID) & 7.80e-6&2.81e-5 & & 8.72e-5& 2.77e-4& & \\ \cline{2-9}
\hline 
\multirow{4}{*}{$\epsilon=8.0$} &RR &\multicolumn{7}{c}{4.41e-5} \\ \cline{2-9}
&GlobalLaplace &1.54e-5 &9.82e-5 &1.53e-4 &3.31e-4 &1.05e-3 &1.58e-3 &2.99e-3 \\\cline{2-9}
&LocalLaplace (IID) &1.51e-5 &1.06e-4 &1.40e-4 &3.27e-4 &8.52e-4 &1.35e-3 &2.29e-3 \\\cline{2-9}
&\cellcolor[HTML]{A8A8A8}LocalLaplace (Non-IID) &\cellcolor[HTML]{A8A8A8}1.06e-5 &\cellcolor[HTML]{A8A8A8}6.99e-5 &\cellcolor[HTML]{A8A8A8}1.14e-4 &\cellcolor[HTML]{A8A8A8}2.19e-4 &\cellcolor[HTML]{A8A8A8}7.50e-4 &\cellcolor[HTML]{A8A8A8}1.16e-3 &\cellcolor[HTML]{A8A8A8}2.22e-3 \\
&adaptive (IID) &7.01e-6 &4.55e-5 & & 1.37e-4& 4.61e-4& & \\ \cline{2-9}
&adaptive (non-IID) &3.89e-6 &1.36e-5 & & 4.34e-5& 1.38e-4& & \\ \cline{2-9}
\hline 
\bottomrule
\end{tabular}}
\caption{Standard deviation of AUC estimated by different methods (DP budget allocation $\alpha = 0.5$ which means that $\epsilon_{\text{localSum}} = \epsilon_{\text{localP}} = 0.5\epsilon$). IID: assigned samples to clients uniformly. Non-IID: assigned samples to clients based on their prediction scores.}\label{tab:std_auc_w_noisy_P_and_N}
\end{table}

\section{Privacy Analysis of Prediction Scores}
\label{sec:sorting_module} 

In this section, we talk about how clients know the ranking results of their prediction scores and the corresponding privacy issue. 

In the vanilla setting, each client $C_k$ can send its shuffled prediction scores $s^k$ to the server. The server then aggregates all the prediction scores and sort them in an increasing order. Each prediction score $s_i^k$ ($i \in [1, M]$) will be assigned a ranking order $r_i^k$. The instance with the highest order will be assigned $r = M - 1$, and the second highest one will be assigned $M -2$ and so on. The smallest ranking score is $r = 0$. The server sends each ranking order $r_i^k$ back to the corresponding client $C_k$ which owns $s_i^k$. 
It's worth mentioning the prediction scores have to be shared with the server for some tasks. For example, the clients have to report the predicted conversion rate (CVR) of the ad impression to the server in online advertisements. The server \footnote{ad exchange} can later on calculate the bid price by multiplying the predicted CVR with  a constant parameter tuned according to the campaign budget and performance ~\cite{bidding2018}. Then the server can select the advertiser who proposed the highest bid price as the winner to show the ad.

\begin{table*}[!htp]\centering
\setlength{\tabcolsep}{0.5em} 
{\renewcommand{\arraystretch}{1.2}
\begin{tabular}{c|c|c|c}\toprule
Top K & $\#$ Positives in Top K &Precision &Recall \\ \hline
1 &1 &1 &8.52e-6 \\ \hline
5 &4 &0.8 &3.41e-5 \\ \hline
10 &8 &0.8 &6.82e-5 \\ \hline
50 &43 &0.86 &3.67e-4 \\ \hline
100 &79 &0.79 &6.73e-4 \\ \hline
500 &384 &0.768 &3.27e-3 \\ \hline
1,000 &774 &0.774 &6.60e-3 \\ \hline
5,000 &3,914 &0.7828 &0.0334 \\ \hline
10,000 &7,519 &0.7519 &0.0641 \\ \hline
50,000 &30,793 &0.6159 &0.2625 \\ \hline
100,000 &52,268 &0.5227 &0.4455 \\
\bottomrule
\end{tabular}}
\caption{Top-k analysis on Criteo data (small)}\label{tab:top_k_analysis_criteo_data_small}
\end{table*}

 \begin{figure*}[ht!]
  \centering
    \includegraphics[width=0.5\linewidth]{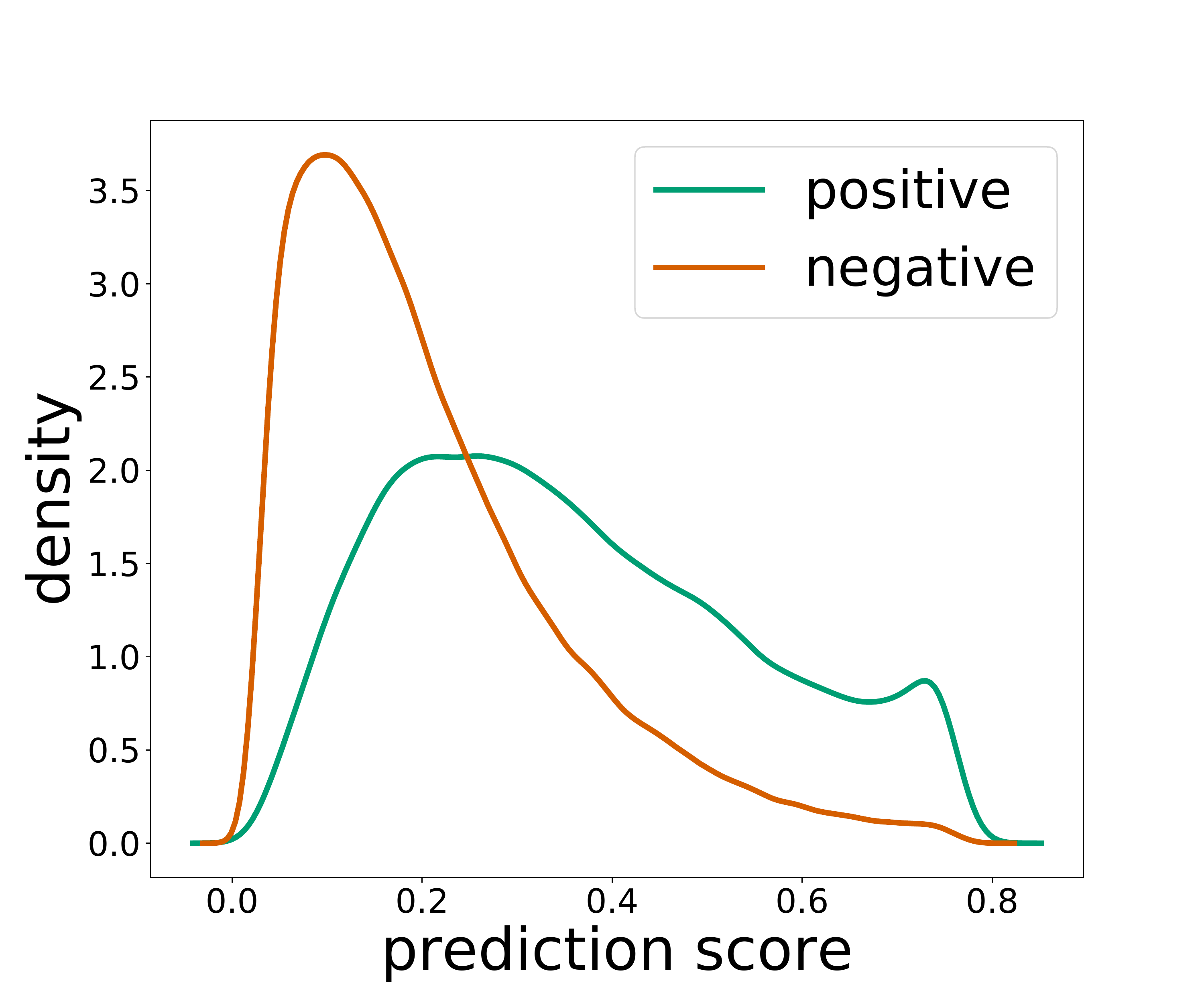}
  \caption{Density of positive and negative instances' prediction scores in Criteo data.}
 \label{fig:criteo_negative_positive_density} 
 \end{figure*}

Interested readers maybe curious is it possible to infer the label information based on the prediction scores. Here we propose a simple attack method by selecting the samples with top-K prediction scores as positive labels. We measure the corresponding guessing performance by precision and recall. As shown in Figure ~\ref{fig:criteo_negative_positive_density} and Table ~\ref{tab:top_k_analysis_criteo_data_small}, positive instances can have a relative higher prediction scores than negative ones at some density areas. For example, we can achieve a $79\%$ precision if we select the instances with top-$100$ prediction scores. However, since the prediction scores $s^k$ are shuffled before sending to the server so that the server has no idea which prediction score belongs to which data sample \footnote{$|s^k| \geq 2$}. 

Instead of the vanilla setting, we also propose some alternative solutions to achieve the sorting goal. We may leverage DP to add noise to the prediction scores. Since the prediction score is the output of a softmax/sigmoid function, the corresponding sensitivity $\Delta = 1$. We can then leverage Gaussian or Laplace mechanism to add noise to the prediction scores. The corresponding results can be seen in Table ~\ref{tab:laplace_dp_for_prediction_score_criteo}. We can observe that the utility of the computed AUC is highly sensitive to the privacy budget. We cannot achieve a reasonable AUC utility with a small $\epsilon$ (i.e. $\epsilon \le 10$). 

\begin{table}[ht!]\centering
\setlength{\tabcolsep}{0.5em} 
{\renewcommand{\arraystretch}{1.2}
\begin{tabular}{c|c|c|c|c}\toprule
&Epoch &0 &1 &2 \\ \hline
&Tensorflow &0.7494 &0.7665 &0.7702 \\ \hline 
&scikit-learn &0.7494 &0.7665 &0.7702 \\ \hline 
\multirow{13}{*}{$\epsilon$} &1 &0.5374 &0.5416 &0.5420 \\ \cline{2-5}
&2 &0.5725 &0.5795 &0.5830 \\ \cline{2-5}
&3 &0.5976 &0.6105 &0.6130 \\ \cline{2-5}
&4 &0.6239 &0.6367 &0.6387 \\ \cline{2-5}
&5 &0.6413 &0.6569 &0.6607 \\ \cline{2-5}
&6 &0.6571 &0.6732 &0.6786 \\ \cline{2-5}
&7 &0.6698 &0.6870 &0.6903 \\ \cline{2-5}
&8 &0.6797 &0.6976 &0.7022 \\ \cline{2-5}
&9 &0.6885 &0.7060 &0.7115 \\ \cline{2-5}
&10 &0.6952 &0.7124 &0.7191 \\ \cline{2-5}
&50 &0.7454 &0.7625 &0.7663 \\ \cline{2-5}
&100 &0.7484 &0.7654 &0.7692 \\ \cline{2-5}
&1000 &0.7494 &0.7665 &0.7702 \\ \cline{2-5}
\hline
\end{tabular}}
\caption{AUC calculated with $\epsilon$-DP for prediction scores with Laplace mechanism on Criteo dataset}\label{tab:laplace_dp_for_prediction_score_criteo}
\end{table}

To achieve both privacy and utility, we can leverage secure multi-party computation (MPC) technique ~\cite{mpc2011,mpc2012} to achieve the sorting goal. For example,  Hamada et al. ~\cite{mpc2012} constructed a quicksort protocol from the quicksort algorithm with their MPC sorting protocol. The resultant protocol can sort $32$-bit words and $1, 000, 000$ secret-shared values in $1, 227$ seconds, while existing sorting protocols cannot sort within $3, 600$ seconds.

In this paper, we adopt the vanilla setting to describe our AUC computation technique by default, since the prediction scores are shared between the server and clients. MPC based sorting protocol can replace our vanilla sorting method if necessary (privacy of prediction scores is a concern). 

\begin{figure*}[ht!]
\captionsetup[subfigure]{labelformat=empty}
  \begin{subfigure}{0.49\linewidth}
  \centering
    \includegraphics[width=1.0\linewidth]{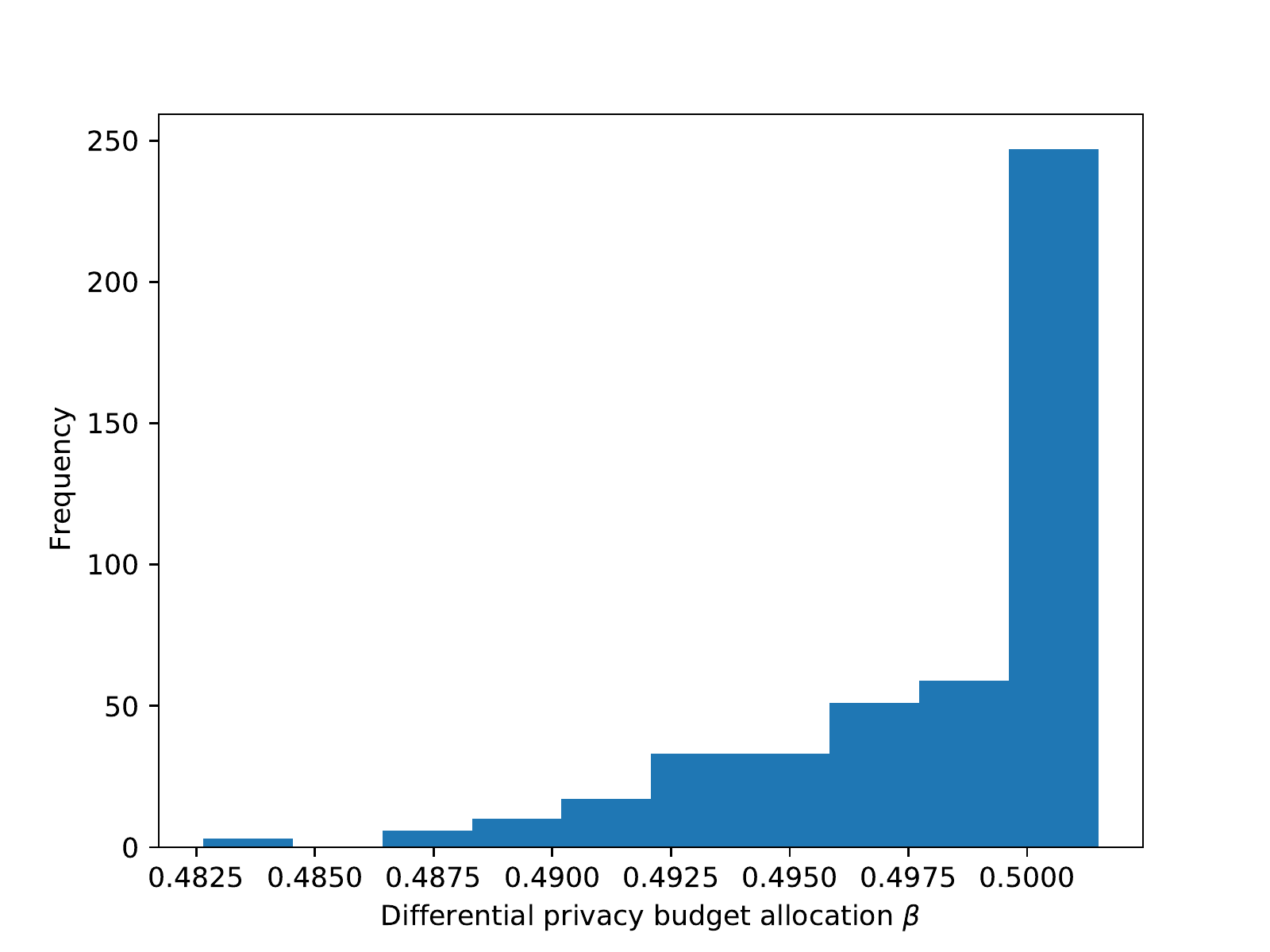}
      \caption{(a): Distribution of local $\beta$ in the IID setting. }
  \end{subfigure}
  \begin{subfigure}{0.49\linewidth}
  \centering
    \includegraphics[width=1.0\linewidth]{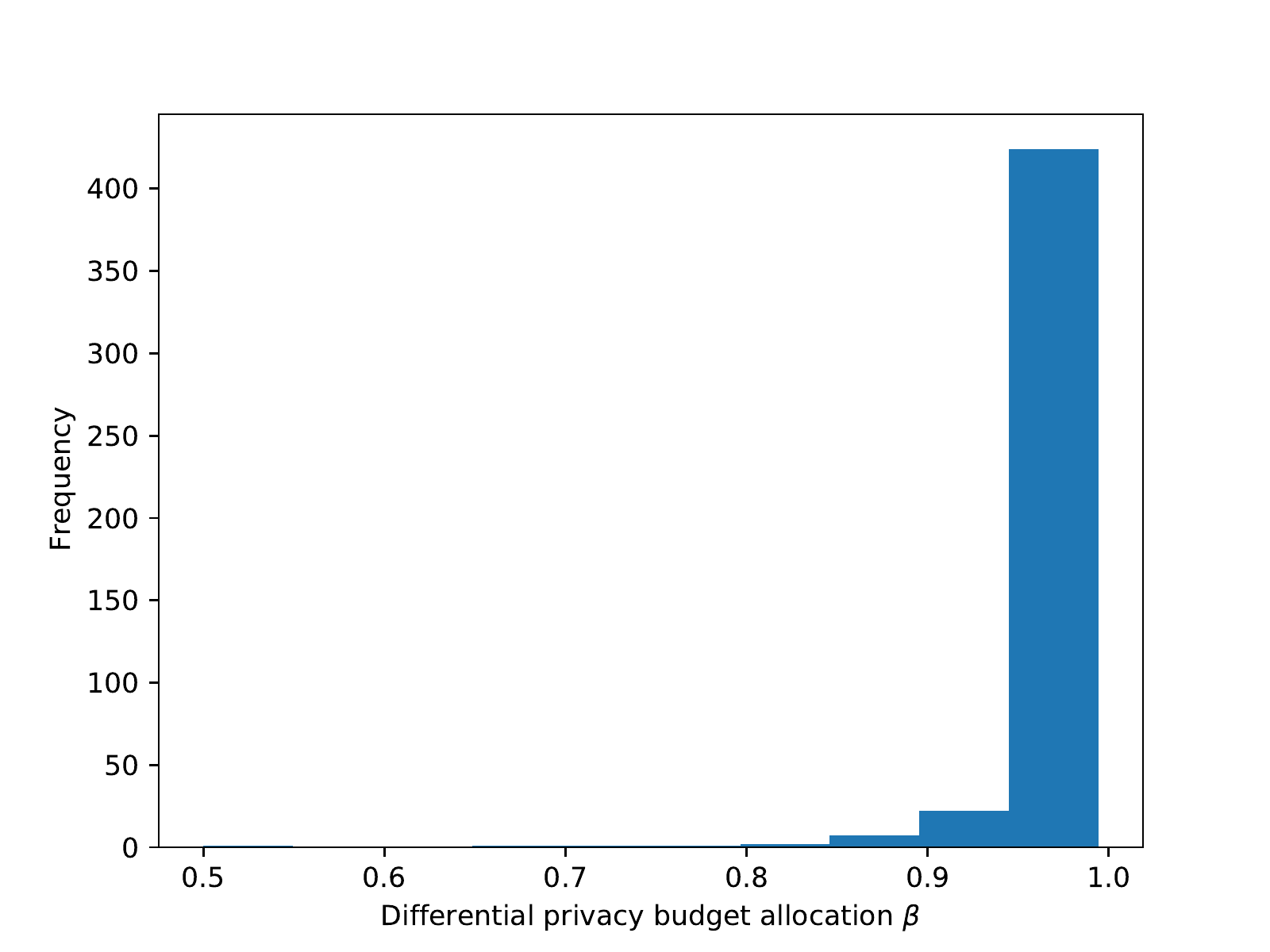}
    \caption{(b): Distribution of local $\beta$ in the Non-IID setting.}
  \end{subfigure}
  
   \caption{Distribution of local $\beta$}
 \label{fig:dist_of_local_beta} 
 \end{figure*}

\section{Figure: ~\ref{fig:dist_of_local_beta}: Distribution of local $\beta$}
\label{sec:dist_of_local_beta}

The distribution of local $\beta$ is shown in Figure ~\ref{fig:dist_of_local_beta}.

\section{Computation resources}
\label{sec:computation_resources}

We conduct our experiments over a Macbook Pro with 2.4 GHz 8-Core Intel Core i9 and 64 GB 2667 MHz DDR4. Each epoch of run of Criteo takes about 2.5 minutes.



\end{document}